\documentclass[10pt,twocolumn,letterpaper]{article}

\usepackage{cvpr}              
\definecolor{cvprblue}{rgb}{0.21,0.49,0.74}
\usepackage[pagebackref,breaklinks,colorlinks,allcolors=cvprblue]{hyperref}
\usepackage{multirow}
\usepackage{graphicx}
\usepackage[inline, shortlabels]{enumitem}
\usepackage{booktabs}
\usepackage{wrapfig}
\usepackage[para,online,flushleft]{threeparttable}
\usepackage[table,RGB]{xcolor}
\newcommand{\STAB}[1]{\begin{tabular}{@{}c@{}}#1\end{tabular}}
\usepackage{diagbox}
\usepackage{marvosym}
\def\paperID{33116}
\def\confName{CVPR}
\def\confYear{2026}

\title{HAD: Heterogeneity-Aware Distillation for Lifelong Heterogeneous Learning}

\author{%
Xuerui Zhang$^{1,3\ast}$, Xuehao Wang$^{2\ast}$, Zhan Zhuang$^{1,4}$, Linglan Zhao$^{5}$\\
Ziyue Li$^{3}$, Xinmin Zhang$^{2}$, Zhihuan Song$^{2}$, Yu Zhang$^{1\textsuperscript{\Letter}}$\\
\small
$^1$Southern University of Science and Technology $^2$Zhejiang University $^3$Technical University of Munich \\
\small
$^4$City University of Hong Kong $^5$Shanghai Jiao Tong University}

\begin{document}

\maketitle

 \renewcommand\thefootnote{} 
\footnotetext{\textsuperscript{\Letter}Corresponding authors. $^{\ast}$Equal contribution.}

\begin{abstract}
Lifelong learning aims to preserve knowledge acquired from previous tasks while incorporating knowledge from a sequence of new tasks. However, most prior work explores only streams of homogeneous tasks (\textit{e.g.}, only classification tasks) and neglects the scenario of learning across heterogeneous tasks that possess different structures of outputs. In this work, we formalize this broader setting as lifelong heterogeneous learning (LHL).
Departing from conventional lifelong learning, the task sequence of LHL spans different task types, and the learner needs to retain heterogeneous knowledge for different output space structures.
To instantiate the LHL, we focus on LHL in the context of dense prediction (LHL4DP), a realistic and challenging scenario.
To this end, we propose the Heterogeneity-Aware Distillation (HAD) method, an exemplar-free approach that preserves previously gained heterogeneous knowledge by self-distillation in each training phase.
The proposed HAD comprises two complementary components, including a distribution-balanced heterogeneity-aware distillation loss to alleviate the global imbalance of prediction distribution and a salience-guided heterogeneity-aware distillation loss that concentrates learning on informative edge pixels extracted with the Sobel operator.
Extensive experiments demonstrate that the proposed HAD method significantly outperforms existing methods in this new scenario.
\end{abstract}    
\section{Introduction}
\label{sec:intro}
Lifelong learning, also known as continual or incremental learning, has garnered significant attention since it holds the potential to continually adapt to a sequence of new tasks from the data stream~\citep{dohare2024loss,lee2024continual, 10896871}.
The primary objective of lifelong learning is to address the catastrophic forgetting problem~\citep{mccloskey1989catastrophic}, which refers to the performance degradation on previously learned tasks after learning new tasks in the absence of historical data.

Existing lifelong learning methods~\citep{zhao2024safe,yang2024domain} in the field of computer vision are primarily designed within the context of homogeneous tasks (\textit{e.g.}, classification-only or segmentation-only tasks), limiting the applicability to broader scenarios.
Specifically, the lifelong learning setting often assumes the arrival of homogeneous tasks, overlooking real-world scenarios where heterogeneous tasks (\textit{e.g.}, a mixture of classification and regression tasks) emerge continuously.
Moreover, sequentially handling heterogeneous tasks, which requires the integration of heterogeneous knowledge, remains underexplored.
Those limitations present challenges to traditional lifelong learning and necessitate extending lifelong learning to a novel scenario of lifelong heterogeneous learning (LHL), under which the incoming tasks are heterogeneous (\textit{e.g.}, a data stream with a mixture of regression and classification tasks).

\begin{figure*}
    \centering
\includegraphics[width=0.78\linewidth]{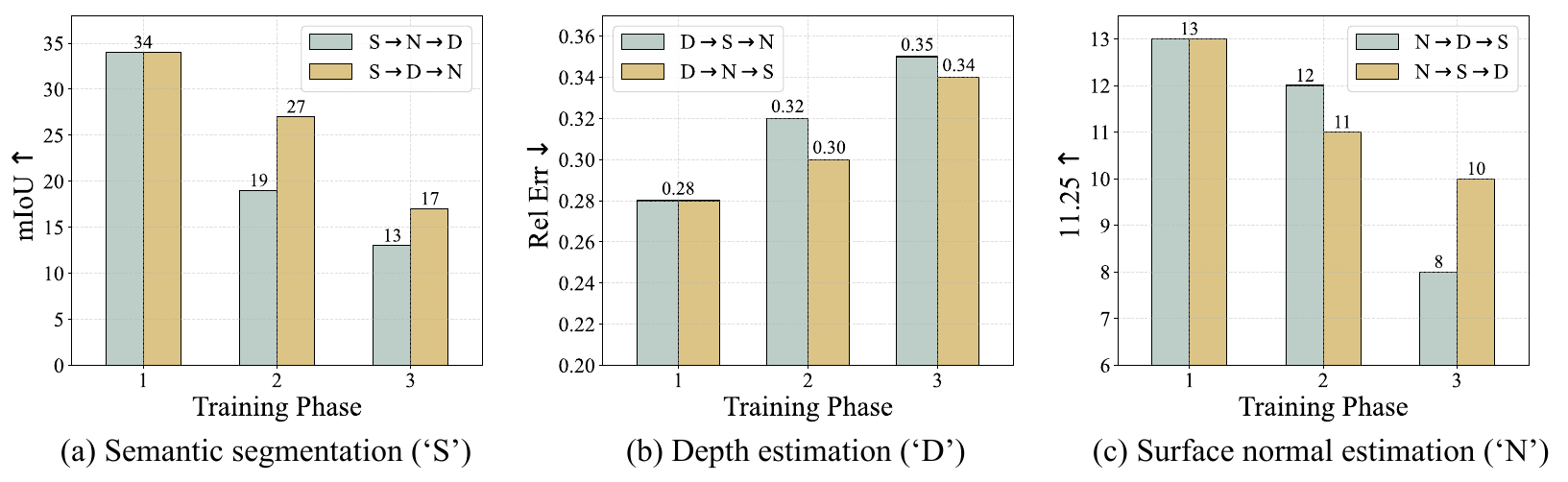}
\vskip -0.1in  
\caption{Vanilla training under LHL4DP. To assess the impact of catastrophic forgetting, we shuffle the learning sequences of three DP tasks. Each figure illustrates how the performance of a given task varies as the training phase proceeds, where the number in the horizontal axis denotes the task index in each sequence of three DP tasks.
The performance metric is indicated above each column.
The symbol $\uparrow$ ($\downarrow$) signifies that a higher (lower) value denotes better performance.}
\label{fig_forgetting}
\vskip -0.2in
\end{figure*}

To instantiate the LHL setting, in this paper, we consider the dense prediction (DP) tasks, a fundamental class of computer vision problems~\cite{zuo2022vision}.
DP tasks are particularly challenging, as these tasks aim to learn a mapping from input images to pixel-wise annotated labels~\citep{kim2023universal}, with heterogeneous label spaces for different DP tasks (\textit{e.g.}, discrete class labels and continuous depth maps).
Recently, the need for the capability of performing multiple DP tasks within a single model has been raised, especially in autonomous driving~\citep{chen2018multi} and intelligent agents~\cite{cai2023open}. Previous works have explored joint training across multiple tasks to enable models to learn these heterogeneous capabilities~\citep{vandenhende2021multi,depthanything,wang2025mtsam}.
However, due to privacy concerns and temporal inconsistency in data collection~\citep{zhou2023deep,yang2024depth, xu2025what}, retaining all previous data and jointly training on multiple tasks as new tasks arise becomes impractical.
Therefore, adopting the LHL setting for DP tasks (LHL4DP) is necessary.
In this scenario, the input data across tasks originates from the same domain, but the tasks encountered sequentially are heterogeneous DP tasks.
Under this scenario, we investigate the presence of the \emph{catastrophic forgetting} phenomenon by performing vanilla training on sequentially coming tasks, with the experimental settings described in Sec.~\ref{sec:exp_setting}.
As shown in Fig.~\ref{fig_forgetting}, all tasks suffer from catastrophic forgetting regardless of the order of coming tasks.

To mitigate this issue under LHL4DP, we propose the Heterogeneity-Aware Distillation (HAD) method. HAD performs self-distillation in an exemplar-free manner, \textit{i.e.}, without storing historical data.
It maintains the heterogeneous knowledge learned from previous tasks by generating pseudo-labels to guide the knowledge retention.
To improve the effectiveness of the pseudo-label guidance, we propose two novel loss functions in the HAD method.
Firstly, a distribution-balanced heterogeneity-aware distillation (DB-HAD) loss is proposed to mitigate imbalanced pseudo-labels in dense prediction tasks~\citep{li2020analyzing,ren2022balanced,zhong2023understanding} by balancing the distribution of different semantic groups. Specifically, we use the geometric mean to smooth the distillation loss within each group,
which reduces the noise in pseudo-labels.
Secondly, a proposed salience-guided heterogeneity-aware distillation (SG-HAD) loss utilizes the Sobel operator~\citep{sobel} to extract the semantic boundaries of predictions, thereby emphasizing the loss of pixels near semantic boundaries to maintain previous knowledge more effectively.

In summary, the contributions of this paper are threefold.
\begin{enumerate*}[a), series = tobecont, font = \itshape]
    \item We introduce a new scenario, lifelong heterogeneous learning (LHL), and emphasize its unique challenges related to heterogeneous tasks and knowledge, in contrast to traditional lifelong learning. In particular, we investigate a concrete case: LHL for dense prediction (LHL4DP).
    \item We propose the HAD method, which consists of two components (DB-HAD and SG-HAD) to address the imbalance issue of the prediction distributions and extract informative areas.
    \item {Comprehensive experiments across diverse datasets and DP tasks in the LHL4DP scenario validate the effectiveness of the proposed HAD approach.}
\end{enumerate*}
\section{Related Work}
\label{sec:related_work}

\noindent{\bf Lifelong learning.}
Lifelong learning, also known as continual or incremental learning, aims to enable models to continually acquire new knowledge from streaming data while mitigating catastrophic forgetting of previously learned knowledge.
Traditional works on lifelong learning can be broadly classified into three categories~\citep{de2021continual}: replay methods, which store exemplars and replay historical data~\citep{icarl,aljundi2019gradient, der};
regularization methods, which introduce additional regularization terms~\citep{ewc,deng2021flattening,sgp,bhattask};
and parameter isolation methods~\citep{de2021continual,chenlearning}, which assign separate model parameters to each new task while masking parameters associated with previous tasks~\citep{fernando2017pathnet,spg}.
While existing works extend lifelong learning to handle heterogeneity in terms of class attributes~\citep{dong2023heterogeneous, goswami2023fecam}, data distribution~\citep{yangdecentralized}, and model structures~\citep{wound2023heterogeneous}, they primarily focus on single-task-type scenarios and ignore the heterogeneity of tasks. In contrast, we turn our attention to a more challenging scenario, where the learning process involves a series of tasks with heterogeneous outputs.

\noindent{\bf Dense prediction.}
Dense prediction (DP) tasks, such as semantic segmentation, depth estimation, and surface normal prediction, are fundamental in computer vision~\citep{CityScapes,vandenhende2021multi,zuo2022vision}.
Those tasks involve per-pixel discrete label or continuous value prediction, requiring fine-grained feature extraction and globally consistent outputs. In general, they pose greater challenges than image-level prediction tasks~\citep{zuo2022vision}.
To endow the model with a diverse set of visual abilities like us humans, which is required by autonomous driving~\citep{chen2018multi} and intelligent agents~\citep{cai2023open}, joint training of DP tasks has been proposed~\citep{unet,deeplabv3++,ranftl2021vision}.
However, practical constraints (\textit{e.g.}, data privacy, limited resources, and sequential data collection~\citep{zhou2023deep,Zhao_2023_CVPR,zhao2024safe}) render joint training across tasks difficult, which motivates the study of DP tasks under the proposed new scenario LHL.

\noindent{\bf Lifelong learning for dense prediction.}
Existing lifelong learning scenarios for DP tasks are typically tailored for the specific tasks, including incremental depth estimation (IDE) and continual semantic segmentation (CSS). The former focuses on enabling continuous depth estimation in emerging domains~\citep{yang2024domain}, while the latter is concerned with segmentation in incremental shift along class and domain directions~\citep{toldo2024learning}.
We differ from these scenarios in two primary aspects.
First, compared with previous task-specific scenarios, LHL4DP not only reduces the risk of overfitting on the specific task~\citep{zhang2021survey}, but holds the potential of learning knowledge from related vision tasks~\citep{vandenhende2021multi}.
Second, while effective within their scope, these task-specific methods rely on homogeneous, task-specific information, making them unsuitable for a sequence of heterogeneous tasks (\textit{i.e.}, LHL4DP). For instance, discrete class labels~\citep{gong2024continual,yin2025beyond} or classification probabilities~\citep{douillard2021plop,toldo2024learning} commonly used in CSS are not available in regression-based tasks.
Similarly, domain-aware solutions in IDE~\citep{hu2023lifelong, yang2024domain} are designed to address challenges such as domain shift and depth spatial variations. Consequently, they are inapplicable to the LHL4DP setting.

\section{Problem Definition}
Conventional lifelong learning assumes all tasks share the same output structure. However, real-world applications such as DP require handling heterogeneous tasks whose outputs include class labels, depth maps, surface normal vectors, and more.
We formalize this setting as LHL (Sec.~\ref{sec:problem_definition}), extend it to the more realistic scenario LHL4DP (Sec.~\ref{sec:HILDP}), and analyze its challenges (Sec.~\ref{sec:paradigm_analysis}).

\subsection{Lifelong Heterogeneous Learning (LHL)}
\label{sec:problem_definition}

In the LHL setting, let $\mathcal{T}=\{\mathcal{T}_t\}_{t=1}^T$ be a sequence of $T$ heterogeneous tasks \textcolor{blue}{}, where all tasks share a common input space $\mathcal{X}$ but each task $\mathcal{T}_t$ has its own output space $\mathcal{Y}_t$ as
\begin{equation}
\mathcal{T}_t = \{(\mathcal{X}, \mathcal{Y}_t)\}, \qquad t=1,\ldots,T.
\label{eq_1}
\end{equation}
{Due to the heterogeneity between tasks, the output space of each task varies, \textit{e.g.}, continuous or discrete outputs.}
Each task $\mathcal{T}_t$ has its corresponding training dataset $\mathcal{D}_t = \{(x,y)|x \in \mathcal{X}, y \in \mathcal{Y}_t\}$, where $(x, y)$ refers to the input and its corresponding label, respectively.
Here, we assume that training instances in different tasks have no overlap. 
Note that during the $t$-th training phase, only the corresponding dataset $\mathcal{D}_t$ of task $\mathcal{T}_t$ is available, and the datasets of other tasks are unavailable.

The goal of LHL is to design a unified lifelong heterogeneous learner, $\mathcal{F}: \mathcal{X} \rightarrow \bigcup_{i=1}^{T} \mathcal{Y}_i$, capable of continually adapting to the sequence of tasks $\mathcal{T}$.
Specifically, at the $t$-th training phase, the learner is expected to accurately predict outputs over the cumulative heterogeneous tasks $\mathcal{T}_{1:t}$.
This requires the learner to retain knowledge acquired during the previous training phases without access to prior data.

\subsection{LHL for Dense Prediction (LHL4DP)}
\label{sec:HILDP}
The above problem setup is universal and holds the potential to benefit a wide range of downstream heterogeneous tasks.
In this paper, we focus on a challenging and realistic scenario involving a sequence of heterogeneous DP tasks.
Each task $\mathcal{T}_t$ corresponds to a distinct DP task (\textit{e.g.}, semantic segmentation, depth estimation, or surface normal prediction) with a unique output space (\textit{e.g.}, class labels, depth maps, or surface normal vectors).

Formally, the input and output spaces of task $\mathcal{T}_t$ are defined as
\begin{equation}
\mathcal{T}_t = \{(\mathcal{X}, \mathcal{Y}_t)| \mathcal{X} \subseteq \mathbb{R}^{C^{\mathrm{in}} \times H \times W}, \mathcal{Y}_t \subseteq \mathbb{R}^{C^{\mathrm{out}}_t \times H \times W}\},
\end{equation}
where $H$, $W$ denotes the spatial dimensions (\textit{i.e.}, the height and width of input images), $C^{\mathrm{in}}$ denotes the number of input channels
(\textit{e.g.}, $3$ for RGB images), 
$C^{\mathrm{out}}_t$ denotes the number of output channels {for the $t$-th} task $\mathcal{T}_t$, which varies across tasks
(\textit{e.g.}, $1$ for the depth estimation task or $3$ for the surface normal estimation task).
Moreover, $\mathcal{Y}_t$ is continuous for some tasks (\textit{e.g.}, depth estimation and surface normal estimation) or discrete for some tasks (\textit{e.g.}, semantic segmentation), while all tasks share the same input domain.
A detailed comparison between the LHL4DP and existing lifelong learning subcategories is provided in Appendix~\ref{app_comparison}.

\subsection{Challenges}
\label{sec:paradigm_analysis}

LHL4DP poses challenges that extend beyond conventional lifelong learning.
It involves sequentially learning \textit{heterogeneous tasks} with distinct objectives and outputs, complicating the learning process.
These tasks rely on \textit{heterogeneous knowledge} (\textit{e.g.}, 3D scene understanding in depth estimation versus semantic structure in segmentation), making it difficult to balance knowledge retention and forgetting.
Furthermore, the pixel-level nature of DP requires preserving \textit{fine-grained information} while maintaining globally consistent outputs.
Additional discussion of these challenges posed by LHL4DP is provided in Appendix~\ref{app_challenges}.

According to the above analysis, the challenges in the LHL4DP scenario can be attributed to the unique nature of task heterogeneity and further compounded by the added complexity of DP tasks. In the next section, we propose a method to handle those challenges.

\begin{figure*}
    \centering
    \includegraphics[width=0.85\linewidth]{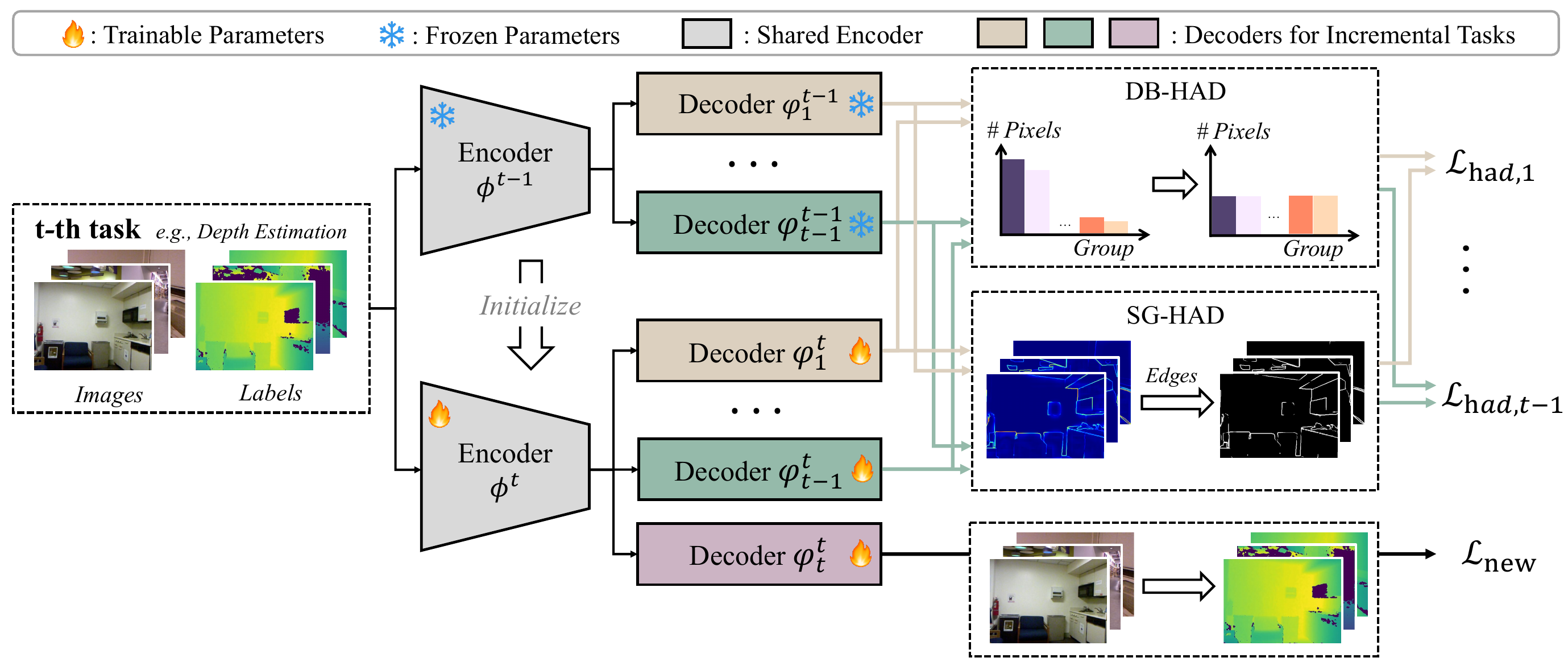}
    \vskip -0.15in
    \caption{The training pipeline of the proposed HAD method in the $t$-th training phase. The HAD method uses the distribution-balanced and salience-guided distillation loss to mitigate forgetting of previous tasks $\mathcal{T}_j\ (j<t)$, all of which are calculated on the pseudo-labels generated by the frozen teacher model $\mathcal{F}^{t-1}_{j} $. Adapting to the new task $\mathcal{T}_t$ is achieved by the task-specific loss function $\mathcal{L}_{\mathrm{new}}$.}
    \label{fig_overview}
    \vskip -0.2in
\end{figure*}
\section{Methodology}
\label{sec_allmethod}
In this section, we introduce the proposed HAD method for the LHL4DP setting.

\subsection{Overview}
\label{sec_overview}

\noindent{\bf Architecture.}
As illustrated in Fig.~\ref{fig_overview},
we employ a task-shared encoder to acquire knowledge from sequential tasks and capture fine-grained image features.
Given the heterogeneity across tasks, task-specific decoders are introduced.
Formally, the learner $\mathcal{F}$ during
the $t$-th training phase comprises: 1) a task-shared encoder
$f_{\phi^t}:\mathcal{X} \rightarrow \mathbb{R}^d $ parameterized by $\phi^t$ that generalizes across tasks; 2) a set of task-specific decoders parameterized by $\{\varphi^t_i\}_{i=1}^t$, where each $g_{\varphi^t_i}:\mathbb{R}^d \rightarrow \mathcal{Y}_i$ maps hidden features to the specific output space of task $\mathcal{T}_i$.
To simplify notation, the prediction function of task $\mathcal{T}_j$ is defined as $\mathcal{F}^t_j(\cdot)=\mathcal{F}(\cdot;\phi^t, \varphi^t_j): \mathcal{X} \to \mathcal{Y}_j$,
and the parameters of learner $\mathcal{F}$ at the training phase $t$ are denoted as $\Phi^t = \{\phi^t, \{\varphi^{t}_j\}_{j=1}^{t}\}$.

\noindent{\bf Heterogeneity-Aware Distillation (HAD).}
To enable continuous knowledge integration from new DP tasks, parameters $\Phi^t$ are initialized from $\Phi^{t-1}$ with the expanded task-specific decoder $\varphi^{t}_{t}$ for task $\mathcal{T}_t$.
Then, $\Phi^t$ is trained on the new DP task $\mathcal{T}_t$ using the task-specific training loss $\mathcal{L}_{\mathrm{new}}$ (\textit{e.g.}, $L_1$ loss for depth estimation task and cross-entropy loss for semantic segmentation tasks) via supervised learning.
Details of $\mathcal{L}_{\mathrm{new}}$ can be found in Appendix~\ref{app_newloss}.

However, as shown in Fig.~\ref{fig_forgetting}, vanilla training on a new task $\mathcal{T}_t$ leads to catastrophic forgetting of previous tasks in the LHL4DP scenario.
A simple remedy is self-distillation (DIS) \citep{pham2022revisiting}, which preserves previous knowledge via the distillation loss $\mathcal{L}_{\mathrm{dis}}$ while learning the new task.
Specifically, during the $t$-th training phase, the previous learners  $\{ \mathcal{F}^{t-1}_i\}_{i=1}^{t-1}$ trained on previous tasks $\mathcal{T}_{1:t-1}$ act as the teacher, while the current learners $\{\mathcal{F}^t_i\}_{i=1}^{t-1}$ being trained on the new task $\mathcal{T}_t$ serve as the student.
To retain prior knowledge of each previous task $\mathcal{T}_j$, we introduce the distillation loss function $\mathcal{L}_{\mathrm{dis}}$ to align the prediction of the student model on $\mathcal{D}_t$ with the pseudo-label generated by the teacher model.
Since datasets $\mathcal{D}_{1:t-1}$ are unavailable and all tasks share a common input space, pseudo-labels for previous tasks can be generated on the dataset $\mathcal{D}_t$ of the new task.
Formally, the total training loss to train $\Phi^t$ is formulated as
\begin{align}
\mathcal{L} = 
&\alpha\underbrace{\sum_{(x,y)\in\mathcal{D}_t} \frac{1}{(t-1)|\mathcal{D}_t|}\sum_j^{t-1} \bar{\mathcal{L}}_{\mathrm{dis},j}(\mathcal{F}_j^t(x),\mathcal{F}_j^{t-1}(x))}_{\mathcal{L}_{\mathrm{dis}}} \notag \\
&+ \underbrace{\frac{1}{|\mathcal{D}_t|} \sum_{(x,y)\in\mathcal{D}_t}\bar{\mathcal{L}}_{t}(\mathcal{F}_t^t(x), y)}_{\mathcal{L}_{\mathrm{new}}},
\label{eq:naive_distillation}
\end{align}
where $\bar{\mathcal{L}}_{\mathrm{dis}, j}$ is the task-specific distillation loss function of task $\mathcal{T}_j$, $\bar{\mathcal{L}}_t$ is the task-specific loss of task $\mathcal{T}_t$, $|\mathcal{D}_t|$ denotes the number of samples in the dataset $\mathcal{D}_t$, and $\alpha$ is the hyperparameter to control the impact of the {distillation loss $\mathcal{L}_{\mathrm{dis}}$}.

However, as illustrated in the next sections, naive distillation in Eq.~(\ref{eq:naive_distillation}) yields limited gains due to imbalanced {pseudo-label} distributions and insufficient focus on salient regions.
To address those issues, we further propose the HAD method, consisting of two loss components for each task $j$: distribution-balanced heterogeneity-aware distillation (DB-HAD) loss $\mathcal{L}_{\mathrm{db}, j}$ and salience-guided heterogeneity-aware distillation (SG-HAD) loss $\mathcal{L}_{\mathrm{sg}, j}$.
Thus, the total training loss to learn $\Phi^t$ in the proposed method is formulated as
\begin{equation}
\mathcal{L} =\frac{\alpha}{2(t-1)} \sum_{j=1}^{t-1}
\mathcal{L}_{\mathrm{had}, j} + \mathcal{L}_{\mathrm{new}},
\label{eq_HISD_obj}
\end{equation}
for each task $\mathcal{T}_{j}$, where the distillation loss $\mathcal{L}_{\mathrm{had}, j}$ is formulated as
\begin{equation}
    \mathcal{L}_{\mathrm{had}, j}=\sum_{(x,y)\in \mathcal{D}_t}\frac{1}{|\mathcal{D}_t|}\big(\mathcal{L}_{\mathrm{db},j}(x) + \mathcal{L}_{\mathrm{sg}, j}(x) \big).
\label{eq_had_j}
\end{equation}
The details of DB-HAD and SG-HAD loss functions are introduced in the following sections.

\subsection{Distribution-Balanced HAD Loss}
\label{sec_part1}
\begin{figure*}[!t]
\centering
    \begin{subfigure}{0.31\textwidth}
        \includegraphics[width=0.9\textwidth]{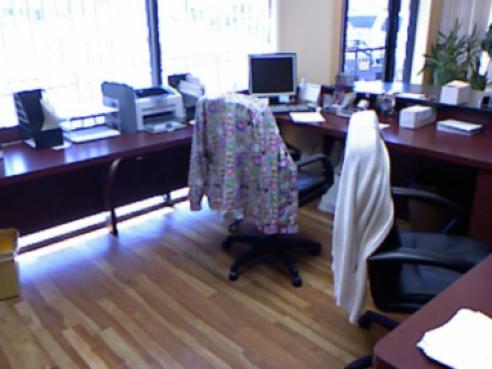}
        \caption{An example of raw images.}
        \label{fig:pdb_img}
    \end{subfigure}
    \begin{subfigure}{0.31\textwidth}
        \includegraphics[width=\textwidth]{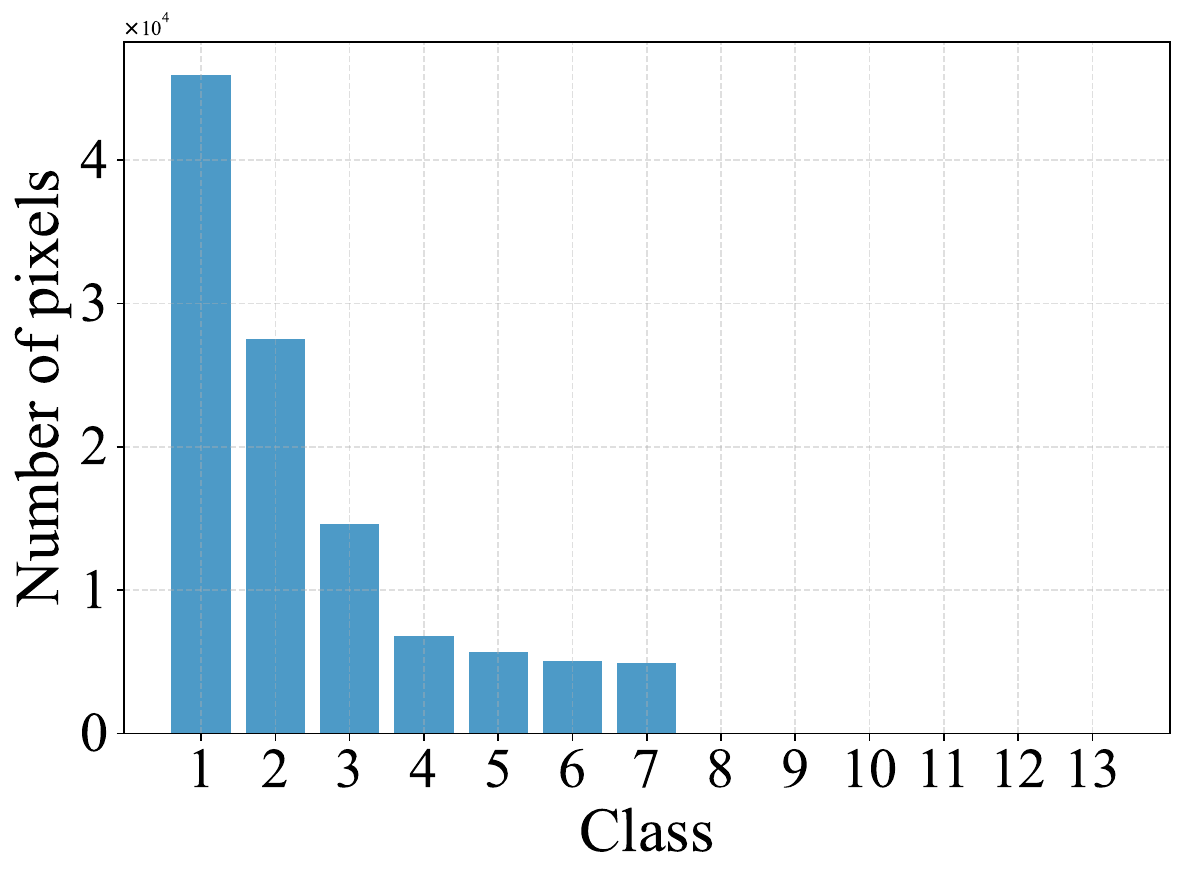}
        \vskip -0.05in
        \caption{Semantic segmentation task.}
        \label{fig:pdb_seg}
    \end{subfigure}
    \begin{subfigure}{0.31\textwidth}
        \includegraphics[width=\textwidth]{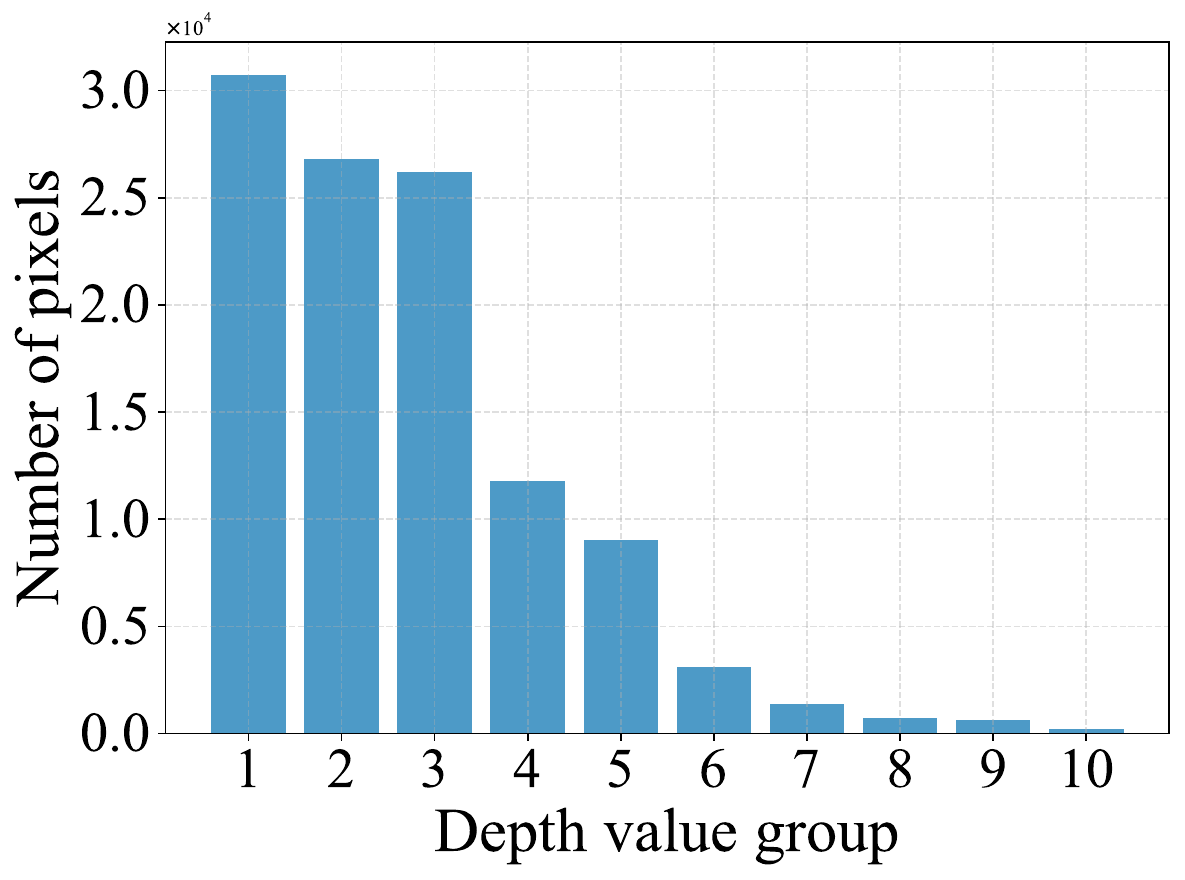}
        \vskip -0.06in
        \caption{Depth estimation task.}
        \label{fig:pdb_dep}
    \end{subfigure}
\vskip -0.13in  
\caption{An illustration of the distribution imbalance in pseudo-labels.
The number of pixels in the semantic segmentation task is counted per class.
In the depth estimation task, we divide the range of pseudo-labels given by the teacher model into ten equal intervals, each of which is a group, and then the ten groups are sorted based on the number of pixels in each group.}
\label{fig_motivation_pdb}
\vskip -0.13in
\end{figure*}

To preserve previous knowledge while learning new tasks, the teacher model generates pseudo-labels on the new training data to revise the heterogeneous knowledge of previous tasks.
However, we observe that the distribution of generated pseudo-labels is imbalanced.
To address the imbalance issue, we propose the distribution-balanced heterogeneity-aware distillation (DB-HAD) loss, which first partitions image pixels into semantic groups and then balances their respective contributions.

{\noindent{\bf Imbalance issue.} 
Generally, DP tasks can be categorized into pixel-level classification and pixel-level regression tasks.
To illustrate the imbalance phenomenon across these two types of DP tasks, Fig.~\ref{fig_motivation_pdb} visualizes the distribution of pseudo-labels generated by the learner after the first training phase on raw images from the new task data, exhibiting an imbalance pixel-wise distribution of class labels for the classification task (\textit{e.g.}, semantic segmentation in Fig.~\ref{fig:pdb_seg}) and values for the regression task (\textit{e.g.}, depth estimation in Fig.~\ref{fig:pdb_dep}).
This phenomenon is widespread across different tasks rather than being limited to our experiments~\citep{foregroundbackground}, posing a risk to effective knowledge retention during the learning phase of the new task~\citep{jiao2018look}.

\noindent{\bf Group partition.} 
A group is defined as a collection of pixels that share similar semantics, as decided by their pseudo-labels.
For each task $\mathcal{T}_j$, the generated pixel-level pseudo-labels $\mathcal{F}^{t-1}_j(x) \in \mathbb{R}^{H \times W}$ on the image $x \in \mathcal{D}_t$ are divided into $C_j$ non-overlapping groups.
For a pixel-level classification task $\mathcal{T}_j$ in DP, a group corresponds to a class. The number of groups $C_j$ equals the number of classes.
Concretely, for each class $c \in \{i\}_{i=1}^{C_j}$, we construct binary masks $M^x_{c,j} \in \{0, 1\}^{H \times W}$ that indicates the presence of class $c$: $M^x_{c,j}[m,n] = \mathbb{I}(\mathcal{F}^{t-1}_j(x)[m, n] = c)$, where $\mathbb{I}(\cdot)$ is the indicator function, $m\in\{1,\dots, H\}$, $n\in\{1,\dots,W\}$ denote the indices of the mask, and $A[m,n]$ for a matrix $A$ denotes the $(m,n)$-th entry in $A$.
For a pixel-level regression task $\mathcal{T}_j$ in DP, we first obtain a scalar value per pixel by averaging across the channel dimension of size $C^{\mathrm{out}}_j$.
The resulting continuous values are then min–max normalized ~\citep{minmax} into the interval [0,1]
and binarized into two groups ($C_j = 2$), \textit{i.e.}, foreground and background~\citep{foregroundbackground}, using a threshold $\tau \in (0,1)$. This yields two masks: $M^x_{1,j}[m,n] = \mathbb{I}(\mathcal{F}^{t-1}_j(x)[m,n] < \tau)$, and $M^x_{2,j}[m,n] = \mathbb{I}(\tau \leq \mathcal{F}^{t-1}_j(x)[m,n])$.

\noindent{\bf Loss function.} 
Inside each group, we compute the \emph{geometric mean} of the per-pixel distillation loss to mitigate the impact of inaccuracies and noise~\citep{tao2008geometric}. The group losses are then averaged arithmetically, ensuring that each group contributes equally to the training objective.
Given an input image $x$, the loss function $\mathcal{L}_{\mathrm{db}, j}$ of DB-HAD for each task $\mathcal{T}_j$ during the $t$-th training phase can be expressed as
\begin{align}
\label{eq:db}
    \mathcal{L}_{\mathrm{db}, j}(x) = \sum_{c=1}^{C_j} \frac{1}{C_j} \bigg( \prod_{(m,n) \in I_{c,j}(x)} \mathcal{L}_{\mathrm{dis}, j} \big(\mathcal{F}^t_j(x), \notag
    \\ \mathcal{F}^{t-1}_j(x) \big)[m,n] \bigg) ^{\frac{1}{|I_{c,j}(x)|}},
\end{align}
where $\mathcal{L}_{\mathrm{dis}, j}$
denotes the per-pixel distillation loss of task $\mathcal{T}_j$, $I_{c,j}(x)=\{(m,n) | M_{c,j}^x[m,n] > 0\}$ is a set of indices that the corresponding pixel belongs to the group $c$ for the image $x$ of task $\mathcal{T}_j$, and $|I_{c,j}(x)|$ denotes the number of elements in $I_{c,j}(x)$.

\subsection{Salience-Guided HAD Loss}
\label{sec_part2}

In DP tasks, a substantial amount of informative signal resides around semantic boundaries or sharp value transitions~\citep{zhu2020edge, zuo2022vision}. Preserving this information in LHL4DP effectively enhances the retention of heterogeneous knowledge.
Hence, we introduce a complementary salience-guided heterogeneity-aware distillation (SG-HAD) loss to ensure that the model retains information in these pixels.

\noindent{\bf Salient-pixel extraction.}
To enhance distillation, the SG-HAD loss focuses on the edges of the pixel-wise loss map,
corresponding to pixels with sharp variations that carry the most informative signals.
Though ground-truth edges are unavailable, an edge set can be obtained by identifying pixels where the value changes significantly between adjacent pixels~\citep{vincent2009descriptive}.
We first calculate a pixel-wise loss map between the frozen teacher model $\mathcal{F}^{t-1}_j$ and the student model $\mathcal{F}^t_j$ as}
\begin{equation}
\mathbf{I}_j(x) = {\mathcal{L}_{\mathrm{dis},j}} (\mathcal{F}^t_j(x),\mathcal{F}^{t-1}_j(x)) \in \mathbb{R}^{H\times W},
\label{eq_calI}
\end{equation}
where $\mathbf{I}_j(x)$ denotes the pixel-wise distillation loss map of input $x$ for task $\mathcal{T}_j$.
To localize sharp spatial transitions in $\mathbf{I}_j(x)$, we apply the Sobel operator~\citep{sobel}, a discrete differentiation operator that approximates the gradients of the image intensity function. With the horizontal convolution kernel defined as $\mathbf{G}_h = [1,2,1]^{\top} [1,0,-1]$, and the vertical convolution kernel defined as $\mathbf{G}_v = [1,0,-1]^{\top} [1,2,1]$, the Sobel operator conduct the gradient approximation as
\begin{equation}
 \mathbf{G}_j (x)= \sqrt{\left(\mathbf{G}_h * \mathbf{I}_j(x) \right )^2 + \left(\mathbf{G}_v * \mathbf{I}_j(x) \right )^2 },
\label{eq_calG}
\end{equation}
where $*$ denotes the convolution operator and the superscript $(\cdot)^2$ denotes the elementwise square operation.
The edge set $\mathbf{P}_j$ is then selected by thresholding the gradient magnitude map $\mathbf{G}_j$ as
\begin{equation}
    \mathbf{P}_j(x) = \{(m,n) \mid  \mathbf{G}_j(x)[m,n] > k \},
    \label{eq_sobel}
\end{equation}
where $k$ is a hyperparameter controlling the necessary gradient intensity to constitute an edge.

\noindent{\bf Loss function.} We accumulate the pixel-wise loss over the extracted salient edge set only:
\begin{equation}
    \mathcal{L}_{\mathrm{sg}, j}(x) = \sum_{ (m,n) \in \mathbf{P}_j(x)} \frac{1}{|\mathbf{P}_j(x)| } \mathbf{I}_j(x)[m,n],
    \label{eq_sg}
\end{equation}
where $|\mathbf{P}_j(x)|$ denotes the number of elements in $\mathbf{P}_j(x)$.

To summarize, by plugging Eqs.~(\ref{eq:db}) and (\ref{eq_sg}) into Eq.~(\ref{eq_had_j}), we obtain the objective function of the proposed HAD method in the $t$-th training phase. 
By balancing group contributions and emphasizing salient boundaries, the proposed HAD method provides an effective defense against forgetting.
\section{Experiments}

\subsection{Experimental Setup}
\label{sec:exp_setting}

\noindent{\bf Benchmarks.}
We empirically evaluate the performance of the proposed method in the LHL4DP scenario under three well-established DP benchmark datasets~\citep{zhang2025bridgenet, wang2025mtsam}, including \textit{CityScapes}~\citep{CityScapes}, \textit{NYUv2}~\citep{NYUv2}, and \textit{Taskonomy}~\citep{taskonomy} with 2, 3, and 10 heterogeneous tasks, respectively.
The task sequences are randomly selected, and the training data is evenly divided across tasks without overlap. Evaluation is performed using the full test set.
During the training phase of each task, the labels of other tasks are inaccessible.
Additional details about benchmark datasets can be found in Appendix~\ref{app_data}.

\noindent{\bf Baseline methods.} In the LHL4DP setup, we compare the proposed method HAD with lifelong learning methods applicable to this scenario. Specifically, baseline methods include EWC~\citep{ewc}, LWF~\citep{lwf}, iCaRL~\citep{icarl}, DER~\citep{der}, SPG~\citep{spg}, and SGP~\citep{sgp}. For the replay-based baseline methods, DER and iCaRL, we store and replay the pixel-wise predictions to ensure fair comparison with HAD.
In addition to these lifelong learning methods, we also establish two extreme baselines for comparison, including
\begin{enumerate*}[a), series = tobecont, font = \itshape]
    \item Vanilla training, which sequentially trains the coming tasks,
    \item Joint training, where all tasks are trained simultaneously using the complete dataset, which serves as the upper bound.
\end{enumerate*}
For fair comparison, grid searches for task-specific hyperparameters are performed. The introduction and hyperparameter details of baseline methods are shown in Appendix~\ref{app_para}.

\noindent{\bf Evaluation metrics.}
Due to the heterogeneity of task outputs, different evaluation metrics are required to assess model performance across tasks.
However, the disparity among these metrics makes it difficult to compare the overall performance using simple averaging.
Thus, we adopt two metrics to evaluate the performance of a lifelong learning method $m$: 
(1) \textbf{Average relative improvement ($\Delta_{v}^m$)} over the vanilla training method $v$ across tasks, defined as
\begin{equation}
    \Delta^m_{v}=\frac{1}{T}\sum\limits_{i=1}^{T}\frac{1}{M_{i }}\sum\limits_{j=1}^{M_{i}}\frac{(-1)^{s_{i, j}}(E_{i, j}^{m}-E_{i, j}^{v})}{E_{i, j}^{v}},
\end{equation}
where $T$ denotes the number of learned tasks, $M_i$ denotes the number of metrics for task $\mathcal{T}_i$, $E_{i,j}^{m}$ and $E_{i, j}^{v}$ denote the performance of the method $m$ and the vanilla training $v$ in terms of the $j$-th metric in task $\mathcal{T}_i$, respectively, and $s_{i, j}$ equals $1$ if the $j$-th metric in task $\mathcal{T}_i$ prefers a higher value for better performance and otherwise is 0;
(2) \textbf{Mean Rank (MR)}, which is the average rank of each method across different tasks.

\noindent{\bf Implementation details.} We adopt the same DeepLabV3+ architecture~\citep{deeplabv3++} for all methods. Specifically, a pre-trained \textit{ResNet-18}~\citep{He_2016_CVPR} with dilated convolutions~\citep{yu2017dilated} is used as the task-shared encoder across all tasks, and task-specific decoders are Atrous Spatial Pyramid Pooling~\citep{deeplabv3++}. 
To further assess the effectiveness of the proposed method, we also conduct experiments using the \textit{ResNet-50}~\citep{He_2016_CVPR} as the encoder.
Details regarding the hyperparameters and task sequences are presented in Appendix~\ref{app_backbone}.
Sensitivity analysis is provided in Appendix~\ref{app_hyper_ablation}.
The source code is available at \href{https://github.com/Deexaa/HAD}{https://github.com/Deexaa/HAD}.

\begin{table*}[!t]
\centering
\caption{Performance on 3 tasks (\textit{i.e.}, 13-class semantic segmentation, depth estimation, and surface normal prediction) after the last training phase of the \textit{NYUv2} dataset across different encoders. The best results are shown in \textbf{bold}. $\uparrow$($\downarrow$) means that the higher (lower) the value, the better the performance.}
\vskip -0.1in
\resizebox{\linewidth}{!}{
    \begin{tabular}{clccccccccccc}
        \toprule
        & \multirow{4}{*}{\textbf{Method}} & \multicolumn{2}{c}{\textbf{Segmentation}} & \multicolumn{2}{c}{\textbf{Depth}} & \multicolumn{5}{c}{\textbf{Surface Normal}} & \multirow{4}{*}{\textbf{$\Delta_{v}^m \uparrow$}} & \multirow{4}{*}{\textbf{MR$\downarrow$}}\\
        \cmidrule(r){3-4} \cmidrule(r){5-6} \cmidrule(r){7-11} & & \multirow{2.5}{*}{\textbf{mIoU${\uparrow}$}} &  \multirow{2.5}{*}{\textbf{Pix Acc$\uparrow$}} &  \multirow{2.5}{*}{\textbf{Abs Err $\downarrow$}} &  \multirow{2.5}{*}{\textbf{Rel Err$\downarrow$}} & \multicolumn{2}{c}{\textbf{Angle Distance}} & \multicolumn{3}{c}{\textbf{Within $t^{\circ}$}} \\ \cmidrule(r){7-8} \cmidrule(r){9-11} & & & & & & \textbf{Mean $\downarrow$} & \textbf{Median $\downarrow$}  & \textbf{11.25 $\uparrow$} & \textbf{22.5 $\uparrow$} & \textbf{30 $\uparrow$}  \\
        \midrule
        \multirow{10}{*}{\STAB{\rotatebox[origin=c]{90}{\textit{ResNet-18}}}} 
        & Vanilla training & 17.49 & 46.81 & 0.9609 & 0.3328 & 32.45 & 26.92 & 20.72 & 42.56 & 54.73 & $+0.00\%$ & 5.67\\
        & Joint training & 41.84 & 66.14 & 0.5793 & 0.2201 & 31.53 & 25.78 & 22.38 & 44.54 & 56.36 & $+40.83\%$ & 1.00\\
        \cmidrule(r){2-13}
        & EWC & 32.17 & 57.21 & 0.9586 & 0.3493 & 37.52 & 33.08 & 13.47 & 33.37 & 45.35 & $+24.39\%$  & 6.33\\
        & iCaRL & 21.78 & 53.00 & 1.3093 & 0.4561 & 33.07 & 27.73 & 19.45 & 41.20 & 53.43 & $-4.82\%$  & 6.67\\
        & LwF & 31.51 & 57.37 & 0.8986 & 0.3345 & 37.06 & 32.09 & 13.89 & 34.66 & 46.81 & $+24.74\%$  & 5.33\\
        & DER & 21.90 & 53.10 & 1.2735 & 0.4422 & 33.09 & 27.74 & 19.36 & 41.18 & 53.43 & $-3.31\%$  & 6.33\\
        & SPG & 18.10 & 48.15 & 0.8801 & 0.3019 &  32.57 &  26.92 &  20.80 &  42.58 &  54.66 & $+4.01\%$  & 4.33\\
        & SGP & 21.34 & 49.75 & 0.9270 & 0.3181 & 32.87 & 27.15 & 19.99 & 42.15 & 54.31 & $+6.53\%$  & 5.00\\
        \cmidrule(r){2-13}
        & HAD & 35.12 & 59.63 & 0.7410 &  0.2641 & 35.32 & 30.55 & 17.23 & 37.26 & 49.12 & $\mathbf{+32.74\%}$  & \textbf{3.67}\\
        \midrule
        \multirow{10}{*}{\STAB{\rotatebox[origin=c]{90}{\textit{ResNet-50}}}} 
        & Vanilla training & 18.26 & 50.85 & 0.8305 & 0.2725 & 28.21 & 21.93 & 26.74 & 50.99 & 63.08 & $+0.00\%$ & 5.67 \\
        & Joint training & 47.78 & 71.03 & 0.4933 & 0.2149 & 28.10 & 22.24 & 25.32 & 50.43 & 62.95 & $+44.36\%$ & 1.33 \\
        \cmidrule(r){2-13}
        & EWC & 36.55 & 61.75 & 0.7321 & 0.2629 & 33.61 & 28.99 & 18.37 & 39.34 & 51.51 & $+31.08\%$ & 5.67\\
        & iCaRL & 28.08 & 57.41 & 0.9877 & 0.3549 & 30.52 & 25.47 & 22.08 & 44.75 & 57.36 & $+7.12\%$ & 6.33\\
        & LwF & 38.06 &  63.77 & 0.6505 & 0.2466 & 31.84 & 26.28 & 20.67 & 43.33 & 56.07 & $+32.94\%$ & \textbf{4.00}\\
        & DER & 27.12 & 58.04 & 0.7383 & 0.2650 & 31.25 & 26.30 & 21.50 & 43.58 & 55.91 & $+17.83\%$ & 5.67\\
        & SPG & 19.77 & 51.40 & 0.7595 & 0.2626 & 28.38 &  22.29 &  25.79 &  50.41 &  62.87 & $+4.07\%$ & 5.00\\
        & SGP & 18.99 & 51.52 & 0.8368 & 0.2764 &  28.27 & 22.52 &  25.79 & 49.91 & 62.50 & $+1.15\%$ & 6.33\\
        \cmidrule(r){2-13}
        & HAD &  38.70 & 63.70 &  0.6294 &  0.2369 & 32.55 & 27.38 & 19.66 & 41.66 & 53.96 & $\mathbf{+35.71\%}$ & \textbf{4.00} \\
        \bottomrule
    \end{tabular}
}
\label{tab:nyuv2}
\vskip -0.05in
\end{table*}

\subsection{Results}
Tab.~\ref{tab:nyuv2} presents the results of the proposed HAD method on the \textit{NYUv2} dataset, evaluated across various architectures.
As can be seen, compared to baseline methods, the proposed HAD method yields superior average performance across tasks.
Moreover, the HAD method consistently enhances results across different architectures, highlighting its robustness and generalizability.

\begin{table*}[!t]
\setlength{\abovecaptionskip}{0pt}
\setlength{\belowcaptionskip}{0pt}
\centering
\vskip 0.1in
\caption{Performance on 10 tasks: semantic segmentation (Seg.), depth estimation (Dep.), surface normal estimation (Normal), edge-2D detection (E.-2D), reshading (Res.), keypoint-2D detection (K.-2D), edge-3D detection (E.-3D), Euclidean distance (E. D.), curvatures (Curv.), and keypoint-3D detection (K.-3D) on the \textit{Taskonomy} dataset. The best results are shown in \textbf{bold}. $\uparrow$($\downarrow$) means that the higher (lower) the value, the better the performance.
}
\vskip 0.1in
\label{tab_tentask}
\resizebox{0.9\linewidth}{!}{
\begin{threeparttable}
{
\begin{tabular}{l|cccccccccccc}
\toprule

\textbf{Method} & \textbf{Seg. $\downarrow$} & \textbf{Dep. $\downarrow$} & \textbf{Normal $\downarrow$} & \textbf{E.-2D $\downarrow$} & \textbf{Res. $\downarrow$} & \textbf{K.-2D $\downarrow$} & \textbf{E.-3D $\downarrow$} & \textbf{E. D. $\downarrow$} & \textbf{Curv. $\downarrow$} & \textbf{K.-3D $\downarrow$} &  $\Delta_{v}^m \uparrow$ & \textbf{MR}$\downarrow$ \\
 \midrule
 Vanilla & 0.7282 & 0.2673 & 0.2869 & 0.1627 & 0.4658 & 0.5143 & 0.5214 & 0.1968 & 1.8655 & 0.3863 & $+0.00\%$ & 6.20 \\ 
 Joint  &  0.1615 & 0.1071 & 0.1281 & 0.1434 & 0.1487 & 0.2969 & 0.3244 & 0.1054 & 1.3501 & 0.3149 & $+44.57\%$ & 1.00\\
 \cmidrule(r){1-13}
 EWC &  0.4701 & 0.2358 & 0.1945 & 0.1692 &  0.3109 & 0.5327 & 0.5525 & 0.2286 & 1.9835 & 0.4347 & $+6.41\%$ & 6.00 \\
 iCaRL & 0.5704 & 0.2247 & 0.2214 & 0.1846 & 0.3386 & 0.5022 & 0.5126 & 0.1918 &  1.7441 & 0.3965 & $+8.47\%$ & 4.20 \\
LwF & 0.6482 & 0.2628 & 0.2484 & 0.1774 & 0.4718 & 0.5298 & 0.5137 &  0.1879 & 1.9212 & 0.4208 & $+0.68\%$& 6.40 \\
DER & 0.5993 & 0.2872 & 0.2315 &  0.1518 & 0.4584 & 0.5068 &  0.4564 & 0.2674 & 1.8488 & 0.3762 & $+1.94\%$ & 4.90 \\
SPG & 0.6768 & 0.2775 & 0.2617 & 0.1707 & 0.4641 & 0.5481 & 0.5418 & 0.2008 & 1.9087 &  0.3537 & $+0.11\%$ & 6.80 \\
SGP &  0.6751 & 0.2856 & 0.2595 & 0.1668 & 0.4608 & 0.5266 & 0.5319 & 0.2009 & 1.8036 & 0.3765 & $+0.79\%$& 6.00 \\
        \cmidrule(r){1-13}
HAD  &  0.5357 &  0.2110 &  0.1868 & 0.1541 & 0.3379 &  0.4955 & 0.5114 & 0.2187 & 1.7530 & 0.4120 & $\mathbf{+10.90\%}$ & \textbf{3.50} \\ 
 \bottomrule
\end{tabular}
}
\end{threeparttable}
}
\end{table*}

We also present the results of the proposed HAD method in the $10$-task scenario on the \textit{Taskonomy} dataset. As shown in Tab.~\ref{tab_tentask}, the HAD method outperforms the baseline methods, achieving the lowest test loss across the largest number of individual tasks, as well as the best average performance across all tasks. These results further demonstrate the effectiveness of the proposed HAD method.

\begin{figure*}[htb]
    \centering
    \includegraphics[width=0.85\linewidth]{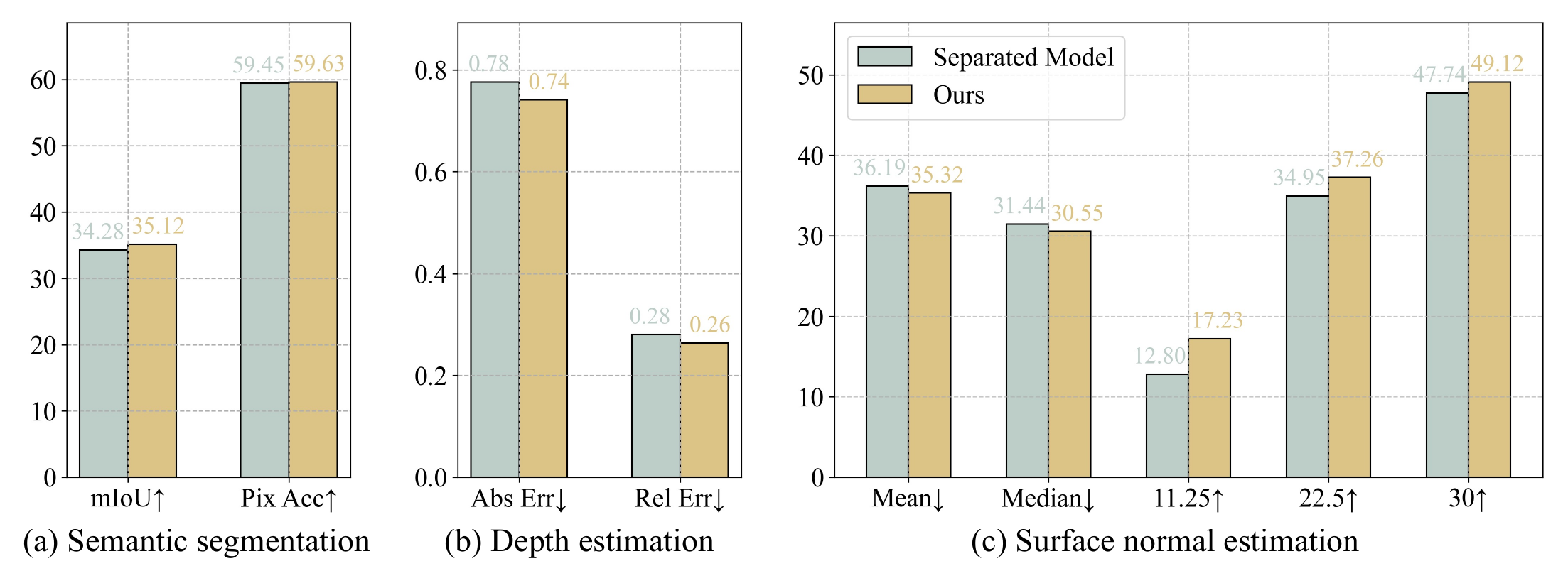}
    \caption{The comparison between training separate models and the proposed HAD. Each figure illustrates the performance improvement of the HAD method in the LHL4DP scenario of a given task. The symbol $\uparrow$ ($\downarrow$) signifies that a higher (lower) value denotes better performance.}
    \label{fig_separate}
\end{figure*}

\begin{minipage}[c]{0.2\textwidth}
\centering
\captionof{table}{Various alternative training paradigms.}
\vskip -0.05in
\label{tab_ab_dis}
\resizebox{\linewidth}{!}{
    \begin{tabular}{l|cc}
        \toprule
        \textbf{Method}      & {$\Delta_{v}^m \uparrow$} & \textbf{MR$\downarrow$} \\
        \midrule
        Paradigm$_1$  &  $-7.76\%$ & 2.00 \\ 
        Paradigm$_2$  & $-6.65\%$  & 3.00 \\ 
        Paradigm$_3$  & $-10.94\%$ & 3.00 \\
        \midrule
        DIS  & $+0.00\%$  & 1.33 \\ 
        \bottomrule     
    \end{tabular}
}
\end{minipage}
\begin{minipage}[c]{0.23\textwidth}
\centering
\captionof{table}{Ablation study.}
\vskip -0.15in
\label{tab_ab_all}
\resizebox{\linewidth}{!}{
\begin{tabular}{l|cc}
    \toprule
    \textbf{Method}      & {$\Delta_{v}^m \uparrow$} & \textbf{MR$\downarrow$}\\
    \midrule
    HAD    &    $+0.00\%$   &   1.00  \\
    HAD \textit{w/o} $\mathcal{L}_{\mathrm{sg}}$ & $-1.27\%$ & 3.00 \\
    HAD \textit{w/o} $\mathcal{L}_{\mathrm{db}}$ & $-2.84\%$ & 3.33 \\ 
    \midrule
    HAD \textit{w/} Arithmetic   &  $-6.29\%$   &       4.00  \\  \midrule
    HAD ($\hat{C}_j$=5)  & $-3.64\%$ & 5.00  \\
    HAD ($\hat{C}_j$=10)  & $-5.43\%$ & 5.67 \\
    HAD ($\hat{C}_j$=15)  & $-9.72\%$ & 5.00 \\
    \bottomrule     
\end{tabular}
}
\end{minipage}

\subsection{Ablation study}

In this section, we conduct ablation studies to assess the contribution of every component in HAD.

\noindent{\bf Effectiveness of DIS.} 
We compared the chosen training paradigm DIS of the proposed HAD method with several alternative paradigms used for maintaining previous knowledge in the $t$-th ($t>1$) training phase:
1) While $\{\phi^t, \varphi^t_t\}$ is being trained on $\mathcal{L}_{\mathrm{new}}$, only the decoders $\{\varphi_t^j\}_{j=1}^{t-1}$ for the old tasks are updated using pseudo-labels.
2) After training $\{\phi^t, \varphi^t_t\}$ on $\mathcal{L}_{\mathrm{new}}$, the decoders $\{\varphi_t^j\}_{j=1}^{t-1}$ for the old tasks are updated using pseudo-labels.
3) We only update the new decoder $\varphi_t^t$ with $\mathcal{L}_{\mathrm{new}}$.
All the training paradigms update the encoder and decoder of the first task on $\mathcal{L}_{\mathrm{new}}$.
As shown in Tab.~\ref{tab_ab_dis}, HAD consistently outperforms these alternatives.

\noindent{\bf Effectiveness of DB-HAD.}
We assess the effectiveness of the DB-HAD from three perspectives.
First, as depicted in Tab.~\ref{tab_ab_all}, removing DB-HAD from the baseline degrades the performance by $2.84\%$.
Second, substituting the geometric mean in Eq.~(\ref{eq:db}) with the arithmetic mean (``HAD \textit{w/} Arithmetic'' in Tab.~\ref{tab_ab_all}) degrades the performance, highlighting the advancement of the geometric mean in mitigating the impact of inaccuracies and noise. 
Third, for regression tasks, we compare our group partitioning with baselines that divide the value range into equally sized intervals for group numbers $\hat{C}_j \in \{5, 10, 15\}$, where each interval has the same width of $\frac{1}{\hat{C}_j}$.
As shown in Tab.~\ref{tab_ab_all}, the proposed HAD method outperforms all variants consistently.

\begin{figure}[!t]
    \centering
    \includegraphics[width=\linewidth]{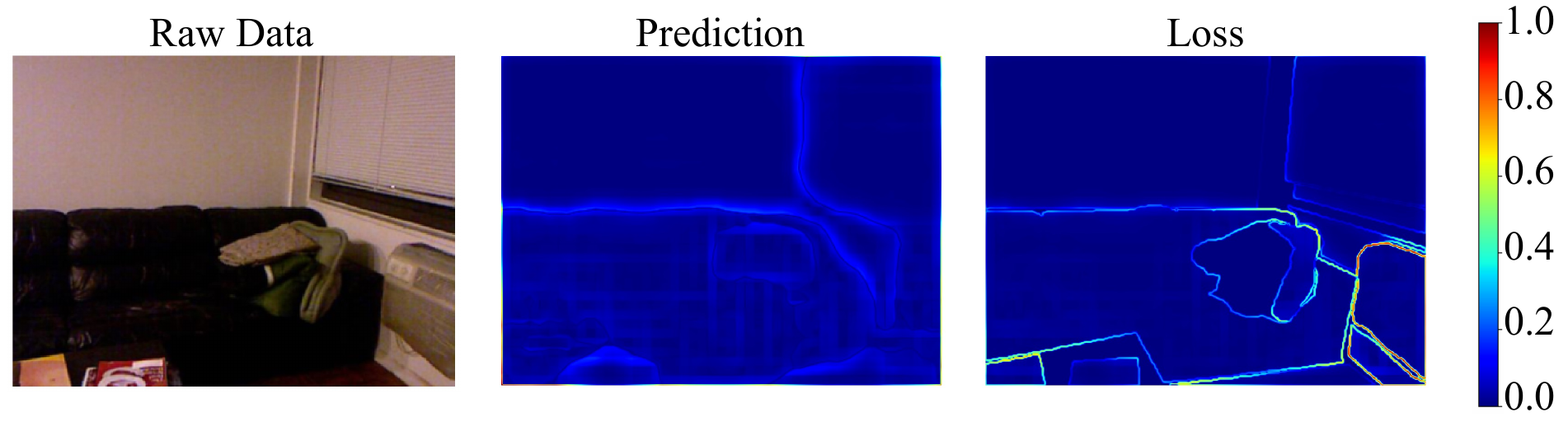}
    \vskip -0.15in
    \caption{Visualization of the raw data (left), the gradient magnitude map of its prediction (middle), and the gradient magnitude map of the loss map (right).}
    \vskip -0.2in
    \label{fig_abvis}
\end{figure}

\noindent{\bf Effectiveness of SG-HAD.} 
As shown in Tab.~\ref{tab_ab_all}, removing the salience-guided loss reduces the performance by $1.27\%$, underscoring the importance of edge-aware focus to HAD.
To assess the effectiveness of the Sobel operator applied to the loss map in Eqs.~(\ref{eq_calI}) and (\ref{eq_calG}), we compare its gradient magnitude map with the one derived from the prediction map.
As shown in Fig.~\ref{fig_abvis}, the gradient magnitude map of the loss map (right) produces more distinct edges than the prediction map (middle), making it better suited for identifying edge sets.

\subsection{Advantages of the LHL4DP Scenario}
\label{sec_scenariobenefits}
We demonstrate the advantages of the LHL4DP scenario by comparing it with training separate task-specific models under the same amount of labeled data. In addition to its practical benefit of enabling the simultaneous support of heterogeneous outputs, LHL4DP offers two key advantages: reduced memory overhead and potential to improve the performance of previously learned tasks.

\noindent{\bf Lower memory overhead.}
Compared with using a shared encoder in the LHL4DP scenario, training separate task-specific models introduces more parameter overhead of $172.23\%$ for additional encoders when using \textit{ResNet-18} with ten tasks.
Notably, this overhead grows with both the encoder complexity and the number of tasks, which is efficiently avoided by the proposed LHL4DP scenario.

\noindent{\bf Better performance.} The comparison on the \textit{NYUv2} dataset is shown in Fig.~\ref{fig_separate}. As can be seen, LHL4DP achieves average improvements of 1.38$\%$, 5.2$\%$, and 9.87$\%$ on the semantic segmentation, depth estimation, and surface normal estimation tasks, respectively, compared to task-specific models.

\emph{Due to page limit, additional experiments, including results for different task sequences, additional model architectures, and on other datasets, are put in Appendix \ref{app_results}.}
\section{Conclusion}
In this paper, we propose a novel lifelong learning scenario named lifelong heterogeneous learning (LHL), which brings unique challenges for traditional lifelong learning. Specifically, we focus on the practical and challenging dense prediction tasks within the LHL scenario (LHL4DP).
To address the unique challenges of LHL4DP, we propose the heterogeneity-aware distillation (HAD) method, which is composed of a prediction distribution balance and a salience-guided distillation loss function. 
The comprehensive experimental results and ablation studies demonstrate the effectiveness of the proposed HAD method. Additional analysis experiments demonstrate the advantages of the proposed LHL4DP scenario.

\section*{Acknowledgments}

This work was supported by National Natural Science Foundation of China under Grant no. 62136005 and Shenzhen fundamental research program JCYJ20250604144724032.

{
    \small
   
    \bibliographystyle{ieeenat_fullname}
    \bibliography{main}
}
\clearpage
\setcounter{page}{1}
\maketitlesupplementary

\appendix
\section{Comparison with Existing Lifelong Learning Scenarios}
\label{app_comparison}

\begin{figure*}[h]
\centering
    \begin{subfigure}{0.4\textwidth}
        \label{fig:cil}
        \includegraphics[width=\textwidth]{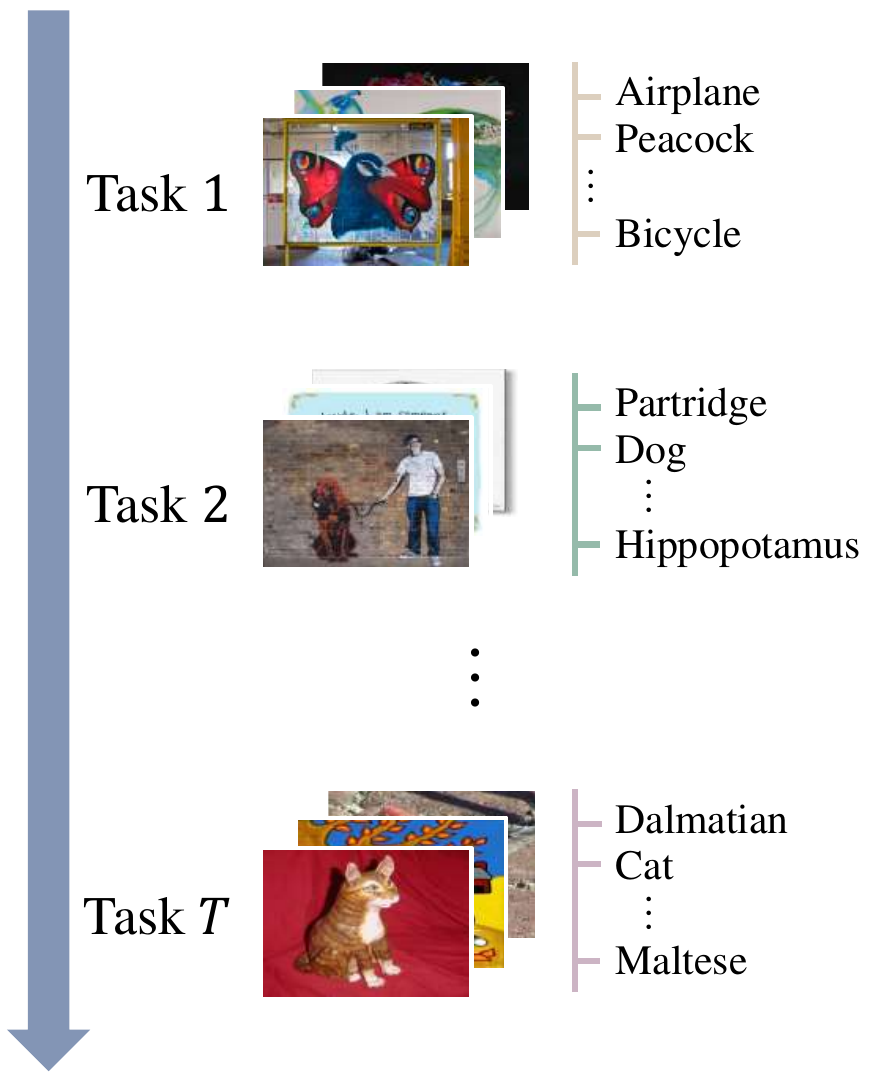}
        \caption{Lifelong Learning.}
    \end{subfigure}
    \begin{subfigure}{0.375\textwidth}
        \label{fig:hil}\includegraphics[width=\textwidth]{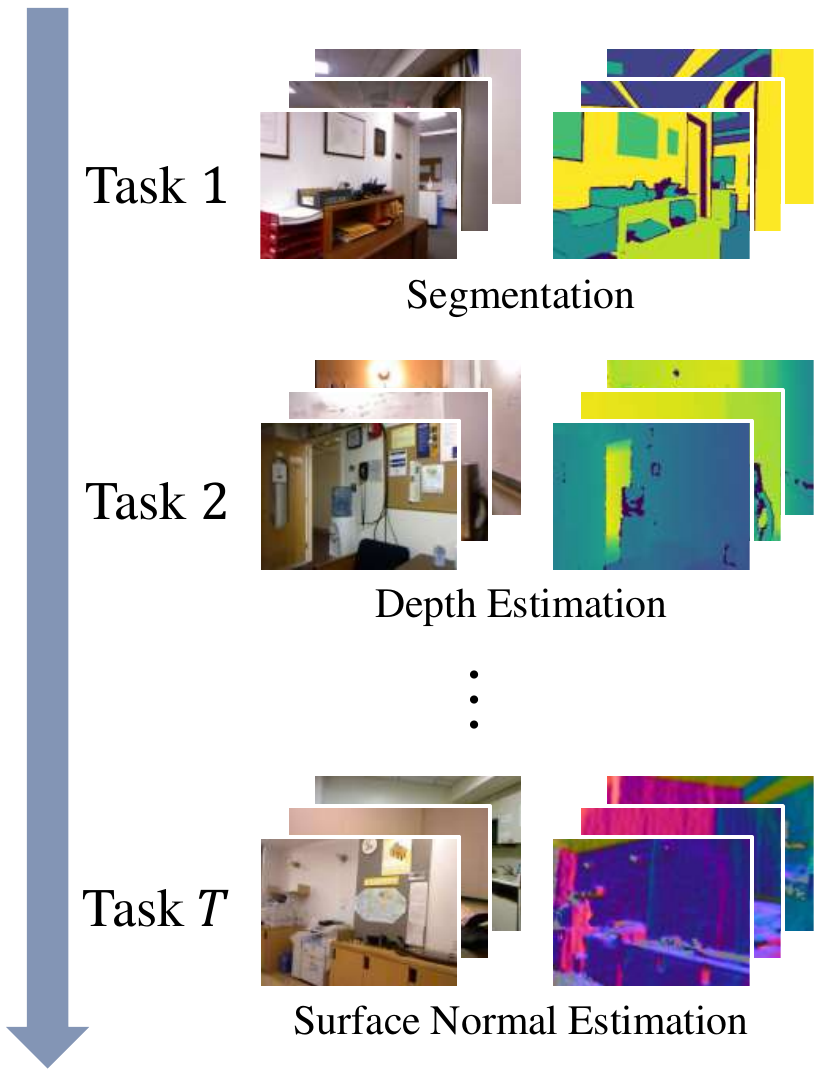}
    \caption{LHL4DP.}
    \end{subfigure}
\caption{Comparison between Lifelong Learning and Lifelong Heterogeneous Learning for Dense Prediction (LHL4DP). (a) CIL incrementally recognizes all encountered classes, while (b) LHL4DP progressively addresses all encountered heterogeneous dense prediction tasks.}
\label{fig_compare}
\end{figure*}

In this section, we analyze the similarities and differences between the proposed lifelong heterogeneous learning for dense prediction (LHL4DP) scenario and the traditional lifelong learning scenario.

\noindent{\bf Similarities.}
The proposed LHL4DP scenario shares three key similarities with the traditional lifelong learning scenario: objectives, settings, and challenges. First, both LHL4DP and lifelong learning aim to achieve performance on sequential tasks comparable to that of joint training across multiple tasks~\citep{wang2024comprehensive}. Second, in both settings, models can be trained for multiple epochs on all data for a given task, while data from previous and future tasks remains inaccessible. Finally, both LHL4DP and lifelong learning face the issue of catastrophic forgetting, where training on the current task leads to the loss of knowledge from previous tasks.

\noindent{\bf Differences.}
Traditional lifelong learning can be classified into three subcategories: class incremental learning (CIL), task incremental learning (TIL), and domain incremental learning (DIL).
Compared with other subcategories, the domain of the training data in DIL varies across tasks, while the number of classes remains consistent across different tasks.
In both CIL and TIL, the domain of the training data remains consistent; however, as the number of tasks increases, so does the total number of classification categories. The key difference between TIL and CIL is that TIL requires a task ID during inference.
However, CIL, TIL, and DIL remain restricted to a single task type, \textit{e.g.}, only classification tasks, and do not support scenarios involving sequentially arriving heterogeneous tasks.
In contrast, the proposed LHL4DP assumes tasks share an input distribution but differ in output types (\textit{e.g.}, class labels and continuous values), which introduces unique challenges for LHL4DP.
Illustration of the comparison is provided in Fig.~\ref{fig_compare}.

\section{Challenges of LHL4DP}
\label{app_challenges}
The unique challenges posed by the proposed LHL4DP scenario are listed as follows.

\noindent{\bf Heterogeneous tasks.}
The technical challenges inherent in LHL4DP are beyond those typically encountered in conventional lifelong learning scenarios.
Different from traditional settings that focus on a single type of tasks (\textit{e.g.}, classification or segmentation), LHL4DP requires learning different types of tasks at different training phases, where each task often involves distinct objective functions and heterogeneous outputs. This results in a more complex and challenging training process.

\noindent{\bf Heterogeneous knowledge.} 
Different tasks require distinct and heterogeneous knowledge representations.
For example, the depth estimation task requires a comprehensive understanding of 3D scenes, while the semantic segmentation task
primarily relies on high-level structured semantic knowledge~\citep{kim2023universal, yin2022transfgu,chen2024neural}.
This divergence presents a challenge for mitigating catastrophic forgetting during the learning of new tasks, highlighting the necessity of strategies that facilitate effective knowledge transfer across heterogeneous tasks.

\noindent{\bf{Fine-grained information.}} 
DP tasks involve producing pixel-level outputs that rely on rich fine-grained information, thereby posing additional challenges~\citep{zuo2022vision}.
This complexity makes retaining previously learned knowledge particularly difficult, requiring strategies capable of preserving fine-grained representations and producing globally coherent outputs across sequential DP tasks.

\section{Experiment Details}
\label{app_alldetails}

\subsection{Details of datasets}
\label{app_data}
To evaluate the performance of the proposed HAD method, we conduct experiments on three datasets with different task numbers as different scenarios: \textit{NYUv2} dataset for 3 tasks, \textit{CityScapes} dataset for 2 tasks, and \textit{Taskonomy} dataset for 10 tasks.

\noindent{\bf\textit{NYUv2} dataset.} This dataset contains 795 training images and 654 testing images in a variety of indoor scenes with ground truth for three tasks (\textit{i.e.}, 13-class semantic segmentation, depth estimation, and surface normal prediction). We use the mean Intersection over Union (mIoU) and Pixel Accuracy (Pix Arr) to evaluate the semantic segmentation task, and use the Absolute Error (Abs Err) and the Real Error (Rel Err) to evaluate the depth prediction task. For the surface normal estimation task, it is evaluated with the mean and the median of angular error measured in degrees, and the percentage of pixels whose angular error is within $11.25$, $22.5$, and $30$ degrees.

\noindent{\bf\textit{CityScapes} dataset.} This dataset comprises 2,975 images for training and an additional 500 images for testing, where we conduct experiments on two tasks (\textit{i.e.}, 7-class semantic segmentation and depth estimation). We use the mean Intersection over Union (mIoU) and Pixel Accuracy (Pix Arr) to evaluate the semantic segmentation task, and use the Absolute Error (Abs Err) and the Real Error (Rel Err) to evaluate the depth prediction task.

\noindent{\bf\textit{Taskonomy} dataset.} We split the 1,390 images from three different views in this dataset into training data for the 10 tasks, reserving one unseen view for testing. The evaluation metric used for performance assessment is the test loss.

\subsection{Heterogeneous Tasks}
\label{app_newloss} 
In the LHL4DP scenario, a new dense prediction task is introduced at each training phase.
To accommodate this sequential heterogeneous setting, we employ a shared encoder across all tasks together with task-specific decoders.
When a new task arrives, a new decoder is instantiated for it, while the decoders learned for previous tasks are retained.
During each training phase, the current task is optimized with its corresponding task-specific loss $\mathcal{L}_{\mathrm{new}}$, whose form depends on the task type.
Tab.~\ref{tab_loss_details} summarizes the heterogeneous tasks in the \textit{NYUv2}, \textit{CityScapes}, and \textit{Taskonomy} datasets, together with their task types and corresponding forms of $\mathcal{L}_{\mathrm{new}}$.

\begin{table*}[htbp]
\centering
\caption{Task-specific forms of $\mathcal{L}_{\mathrm{new}}$ for heterogeneous dense prediction tasks.}
\label{tab_loss_details}
\vskip -0.1in
\small
  \begin{tabular}{c|c|c|c|c}
  \toprule
  Dataset & Task Name & Task Type & Output Channel & Loss Function \\
  \midrule
  \multirow{3}{*}{\textit{NYUv2}}
  & Semantic Segmentation & Classification & 13 & Cross-Entropy Loss \\
  & Depth Estimation & Regression & 1 & $L_1$ Loss \\
  & Surface Normal Prediction & Regression & 3 & Cosine Distance Loss \\
  \midrule
  \multirow{2}{*}{\textit{CityScapes}}
  & Semantic Segmentation & Classification & 7 & Cross-Entropy Loss \\
  & Depth Estimation & Regression & 1 & $L_1$ Loss \\
  \midrule
  \multirow{10}{*}{\textit{Taskonomy}}
  & Semantic Segmentation & Classification & 8 & Cross-Entropy Loss \\
  & Depth Estimation ($z$-buffer) & Regression & 1 & $L_1$ Loss \\
  & Surface Normal Estimation & Regression & 3 & Cosine Distance Loss \\
  & Edge Occlusion Prediction & Regression & 1 & $L_1$ Loss \\
  & Reshading & Regression & 1 & $L_1$ Loss \\
  & Keypoint-2D Prediction & Regression & 1 & $L_1$ Loss \\
  & Edge Texture Prediction & Regression & 1 & $L_1$ Loss \\
  & Euclidean Depth Estimation & Regression & 1 & $L_1$ Loss \\
  & Principal Curvature Estimation & Regression & 2 & $L_1$ Loss \\
  & Keypoint-3D Prediction & Regression & 1 & $L_1$ Loss \\
  \bottomrule
  \end{tabular}
  \end{table*}
\subsection{Baselines}
\label{app_para}

\begin{figure*}[h]
    \centering
    \includegraphics[width=0.9\linewidth]{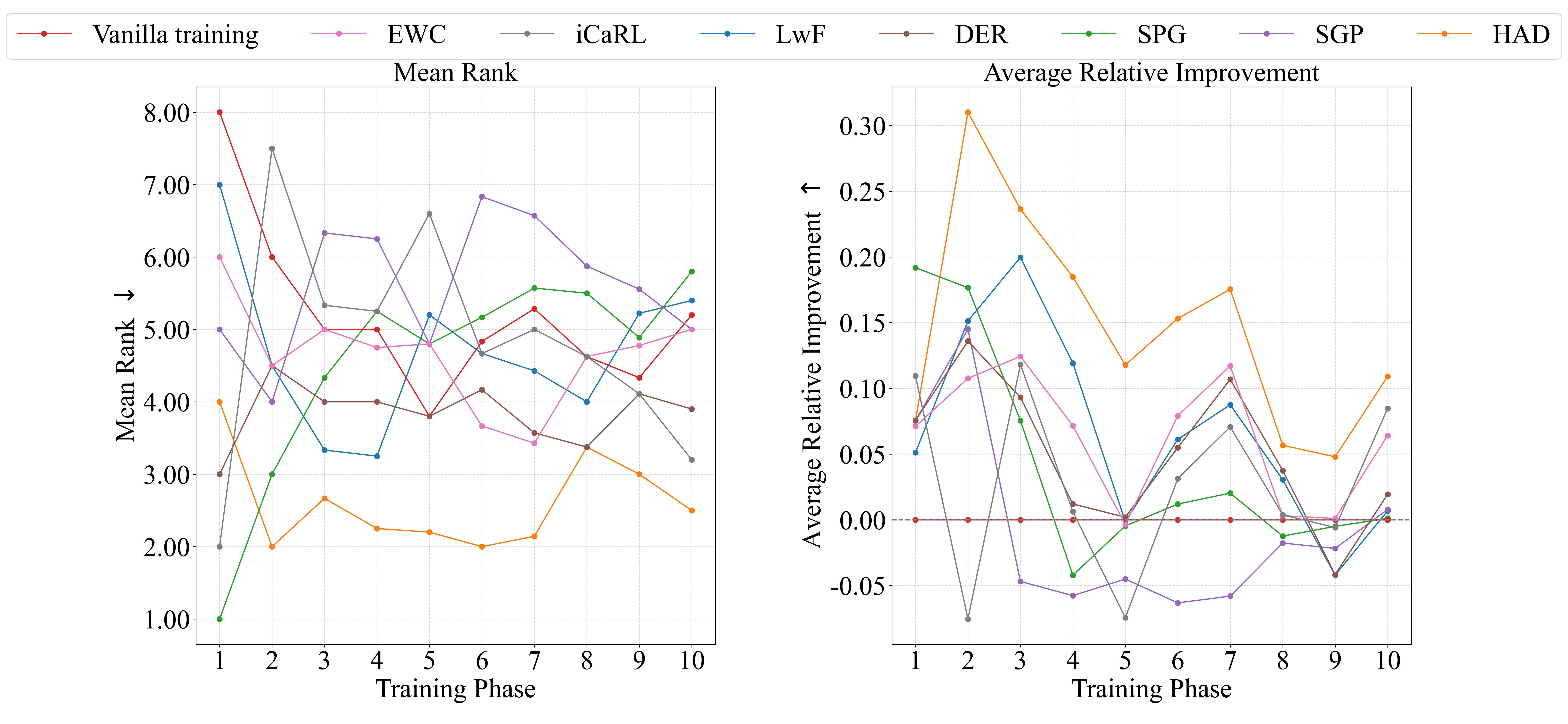}
    \vskip -0.15in
    \caption{Results of different methods during each training phase on the \textit{Taskonomy} dataset.}
    \vskip -0.15in
    \label{app_fig_dynamics}
\end{figure*}

\begin{table}[htbp]
\centering
\caption{Hyperparameter of different methods.}
\label{tab_ab_hyper_baselines}
\vskip -0.1in
\resizebox{\linewidth}{!}{
    \begin{tabular}{c|cc|cc|c}
        \toprule
        \multirow{2}{*}{Method} & \multicolumn{2}{c|}{\textit{NYUv2}} & \multicolumn{2}{c|}{\textit{CityScapes}} &  \textit{Taskonomy} \\
        
        & \textit{Resnet-18} & \textit{Resnet-50} & \textit{Resnet-18} & \textit{Resnet-50} & \textit{Resnet-18}\\ \midrule
        EWC  & $10^{9}$ & $10^{10}$& $10^{6}$ & $10^{3}$ & $10^{9}$\\ 
        iCaRL & 0.01 & 0.1 & 0.01 & 0.1 &  1\\ 
        LWF & 5 & 5 & 0.01 & 0.1 & 0.1\\
        DER  & 0.01 & 0.1 & 1 & 1 & 0.1 \\
        SGP  & 0.1 & 10 & 10 & 1000 & 100\\
        \bottomrule     
    \end{tabular}
}
\end{table}

We compare the proposed HAD method against vanilla training, as well as three categories of traditional IL methods: regularization-based methods including EWC~\citep{ewc}, LWF~\citep{lwf}, and SGP~\citep{sgp}, which constrain the changes in important parameters, representations, and gradients; replay-based methods such as iCaRL~\citep{icarl} and DER~\citep{der}, which store historical data in a fixed-size memory and replay them during the learning of new tasks; and the parameter isolation method SPG~\citep{spg}, which combines orthogonal gradient projections with scaled gradient steps in the important gradient spaces for past tasks. For all replay-based methods, the exemplar size is fixed at 50.

For the hyperparameters of each method, we perform a grid search to select the best-performing configuration. Specifically, we adjust the loss balance weight for LWF, the penalty term for EWC, the distillation loss weights for iCaRL and DER, and the non-negative scale coefficient for SGP. The hyperparameters used for each method across different datasets are summarized in Table~\ref{tab_ab_hyper_baselines}.

\begin{table*}[h]
\centering
\caption{Performance on the \textit{NYUv2} dataset using \textit{Resnet-18} with a shuffled task sequence after the last training phase. The best results for each task are shown in \textbf{bold}. $\uparrow$($\downarrow$) means that the higher (lower) the value, the better the performance.}
\label{tab_seq3}
\resizebox{\linewidth}{!}{
   \begin{tabular}{clcccccccccccc}
        \toprule
        & \multirow{4}{*}{\textbf{Method}} & \multicolumn{2}{c}{\textbf{Segmentation}} & \multicolumn{2}{c}{\textbf{Depth}} & \multicolumn{5}{c}{\textbf{Surface Normal}} & \multirow{4}{*}{\textbf{$\Delta_{b}^T \uparrow$}} & \multirow{4}{*}{\textbf{MR$\downarrow$}}\\
        \cmidrule(r){3-4} \cmidrule(r){5-6} \cmidrule(r){7-11} & & \multirow{2.5}{*}{\textbf{mIoU${\uparrow}$}} &  \multirow{2.5}{*}{\textbf{Pix Acc$\uparrow$}} &  \multirow{2.5}{*}{\textbf{Abs Err $\downarrow$}} &  \multirow{2.5}{*}{\textbf{Rel Err$\downarrow$}} & \multicolumn{2}{c}{\textbf{Angle Distance}} & \multicolumn{3}{c}{\textbf{Within $t^{\circ}$}} \\ \cmidrule(r){7-8} \cmidrule(r){9-11} & & & & & & \textbf{Mean $\downarrow$} & \textbf{Median $\downarrow$}  & \textbf{11.25 $\uparrow$} & \textbf{22.5 $\uparrow$} & \textbf{30 $\uparrow$}  \\
        \midrule
       \multirow{10}{*}{\STAB{\rotatebox[origin=c]{90}{\textit{Sequence 2}}}}
        & Vanilla training & 33.77 & 60.36 & 1.0261 & 0.3592 & 40.76 & 34.74 & 10.16 & 30.98 & 43.17 & $+0.00\%$ & 6.00\\
        & Joint training & 41.84 & 66.14 & 0.5793 & 0.2201 & 31.53 & 25.78 & 22.38 & 44.54 & 56.36 & $+3.09\%$ & 1.00 \\
        \cmidrule(r){2-13}
        & EWC & 29.78 & 56.75 & 0.9217 & 0.3218 & 39.15 & 33.31 & 10.48 & 31.98 & 44.95 & $-0.77\%$ & 5.33\\
        & iCaRL & 23.87 & 53.78 & 1.5976 & 0.5474 & 35.87 & 33.19 & 11.60 & 31.55 & 44.63 & $-27.11\%$ & 7.67 \\
        & LwF & 31.49 & 58.96 & 0.8586 & 0.3044 & 37.66 & 32.54 & 11.76 & 33.50 & 46.17 & $+0.77\%$ & 3.67\\
        & DER & 24.41 & 54.38 & 1.5884 & 0.5428 & 35.78 & 33.00 & 11.71 & 31.84 & 44.96 & $-26.55\%$ & 6.67\\
        & SPG &  34.40 & 60.79 & 1.0025 & 0.3454 & 40.01 & 34.18 & 11.10 & 31.86 & 43.96 & $+0.29\%$ & 4.67 \\
        & SGP & 34.34 &  61.16 & 1.0859 & 0.3803 & 41.30 & 35.83 & 9.62 & 29.30 & 41.47 & $-0.17\%$ & 6.00 \\
        \cmidrule(r){2-13}
        & HAD & 28.86 & 56.83 &  0.7024 &  0.2553 &  35.05 &  30.36 &  12.71 &  36.04 & 49.41 & $\mathbf{+1.18\%}$ & \textbf{3.33} \\
        \midrule
    \multirow{10}{*}{\STAB{\rotatebox[origin=c]{90}{\textit{Sequence 3}}}} 
        & Vanilla training & 22.05 & 51.48 & 1.0131 & 0.3709 & 33.70 & 28.22 & 16.43 & 39.92 & 52.86 & $+0.00\%$ & 5.00\\
        & Joint training & 41.84 & 66.14 & 0.5793 & 0.2201 & 31.53 & 25.78 & 22.38 & 44.54 & 56.36 & $+28.98\%$ & 1.00 \\
        \cmidrule(r){2-13}
        & EWC & 19.60 & 49.40 & 0.9408 & 0.3476 & 38.45 & 34.20 & 11.52 & 31.50 & 43.66 & $+6.62\%$ & 7.33 \\
        & iCaRL & 15.35 & 49.91 & 1.2300 & 0.4093 & 38.33 & 33.66 & 10.53 & 31.16 & 44.15 & $-3.71\%$ & 8.33 \\
        & LwF & 23.46 & 55.26 & 0.8928 & 0.3239 & 37.21 & 32.05 & 16.41 & 35.68 & 47.04 & $+9.43\%$ & 3.67 \\
        & DER & 15.14 & 49.87 & 1.2280 & 0.4076 & 38.17 & 33.48 & 10.63 & 31.41 & 44.41 & $-3.96\%$ & 8.00 \\
        & SPG & 20.98 & 51.08 & 0.9314 & 0.3364 & 34.38 & 28.86 & 15.16 & 38.69 & 51.79 & $+3.10\%$ & 5.00 \\
        & SGP & 22.18 & 52.15 & 0.8878 & 0.3245 & 32.63 & 27.47 & 19.29 & 41.63 & 53.96 & $+2.49\%$ & \textbf{3.00} \\ 
        \cmidrule(r){2-13}
        & HAD & 25.67 & 54.85 & 0.8358 & 0.2992 & 36.96 & 31.81 & 15.81 & 35.83 & 47.39 & $\mathbf{+13.08\%}$ & \textbf{3.00} \\
    \bottomrule
    \end{tabular}
    }
\end{table*}

\subsection{Implementation details}
\label{app_backbone}

\begin{table*}[h]
\centering
\caption{Performance on two tasks after the last training phase (i.e., 7-class semantic segmentation and depth estimation) of the \textit{CityScapes} dataset using \textit{ResNet-18} under two different sequences.}
\resizebox{\linewidth}{!}{
    \begin{tabular}{lcccccccccccc}
        \toprule
        \multirow{4}{*}{\textbf{Method}} & \multicolumn{6}{c}{\textit{Segmentation} $\mathrm{\rightarrow}$ \textit{Depth}} & \multicolumn{6}{c}{\textit{Depth} $\mathrm{\rightarrow}$ \textit{Segmentation}}\\
        \cmidrule(r){2-7} \cmidrule(r){8-13}
        & \multicolumn{2}{c}{\textbf{Segmentation}} & \multicolumn{2}{c}{\textbf{Depth}}  & \multirow{2}{*}{\textbf{$\Delta_{b}^T \uparrow$}} & \multirow{2}{*}{\textbf{MR$\downarrow$}} & \multicolumn{2}{c}{\textbf{Segmentation}} & \multicolumn{2}{c}{\textbf{Depth}}  & \multirow{2}{*}{\textbf{$\Delta_{b}^T \uparrow$}} & \multirow{2}{*}{\textbf{MR$\downarrow$}}\\
        \cmidrule(r){2-3} \cmidrule(r){4-5} \cmidrule(r){8-9} \cmidrule(r){10-11} & \textbf{mIoU${\uparrow}$} &  \textbf{Pix Acc$\uparrow$} &  \textbf{Abs Err $\downarrow$} &  \textbf{Rel Err$\downarrow$} & & & \textbf{mIoU${\uparrow}$} &  \textbf{Pix Acc$\uparrow$} &  \textbf{Abs Err $\downarrow$} & \textbf{Rel Err$\downarrow$} \\
        \midrule
        Vanilla training & 58.40 & 86.84 & 0.0203 & 50.0861 & $+0.00\%$ & 6.50 & 68.44 & 91.45 & 0.0456 & 77.8347 & $+0.00\%$ & 5.50\\
        Joint training & 71.38 & 92.15 & 0.0164 & 43.7236 & $+15.06\%$ & 1.00 & 71.38 & 92.15 & 0.0164 & 43.7236 & $+28.23\%$ & 1.00 \\
        \midrule
        EWC & 66.14 & 90.01 & 0.0203 & 56.8526 & $+0.85\%$ & 6.50 & 65.56 & 90.06 & 0.0217 & 51.9560 & $+19.98\%$ & 5.50 \\
        iCaRL & 67.10 & 91.09 & 0.0204 & 50.2167 & $+4.76\%$ & 5.50 & 57.24 & 87.82 & 0.0221 & 58.8481 & $+13.90\%$ & 7.00 \\
        LwF & 62.67 & 87.98 & 0.0202 &  47.6400 & $+3.50\%$ & 4.50 & 67.52 & 89.54 & 0.0192 & 47.2005 & $+24.91\%$ & 5.00 \\
        DER & 67.15 & 91.11 & 0.0206 & 51.3117 & $+3.99\%$ & 5.50 & 69.20 &  91.90 & 0.0256 & 54.3059 & $+18.92\%$ & \textbf{3.50} \\
        SPG & 68.07 &  91.31 & 0.0484 & 94.8094 & $-51.50\%$ & 5.50 &  69.68 & 91.75 & 0.0418 & 104.0395 & $-5.80\%$ & 5.00 \\
        SGP & 53.00 & 82.38 & 0.0202 & 49.2638 & $-3.06\%$ & 6.00 & 68.07 & 91.31 & 0.0484 & 94.8094 & $-7.16\%$ & 7.50 \\
        \midrule
        HAD &  68.28 & 90.76 &  0.0192 & 51.7254 & $\mathbf{+5.89\%}$ & \textbf{3.50} & 69.12 & 91.12 &  0.0186 &  45.5291 & $\mathbf{+26.70\%}$ & \textbf{3.50} \\
        \bottomrule
    \end{tabular}
}
\label{tab:city}
\end{table*}

\begin{table*}[h]
\centering
\caption{Performance on two tasks after the last training phase (i.e., 7-class semantic segmentation and depth estimation) of the \textit{CityScapes} dataset using \textit{Resnet-50}. The best results for each task are shown in \textbf{bold}. $\uparrow$($\downarrow$) means that the higher (lower) the value, the better the performance. }
\vskip -0.1in
\resizebox{0.64\linewidth}{!}{
    \begin{tabular}{lcccccc}
        \toprule
        \multirow{2}{*}{\textbf{Method}} 
        & \multicolumn{2}{c}{\textbf{Segmentation}} & \multicolumn{2}{c}{\textbf{Depth}}  & \multirow{2}{*}{\textbf{$\Delta_{b}^T \uparrow$}} & \multirow{2}{*}{\textbf{MR$\downarrow$}}\\
        \cmidrule(r){2-3} \cmidrule(r){4-5}
        & \textbf{mIoU${\uparrow}$} &  \textbf{Pix Acc$\uparrow$} &  \textbf{Abs Err $\downarrow$} &  \textbf{Rel Err$\downarrow$}  \\
        \midrule
        Vanilla training &  64.93 & 88.93 & 0.0168 & 40.0604 & $+0.00\%$ & 5.50  \\
        Joint training &  76.49 & 93.91 & 0.0155 & 45.7162 & $+4.26\%$ & 3.00  \\
        \midrule
        EWC &  69.44 & 90.60 & 0.0154 & 41.4725 & $+3.41\%$ & \textbf{3.50} \\
        iCaRL & 72.80 & 92.60 & 0.0166 & 47.3969 & $-0.22\%$ & 7.00  \\
        LwF &  76.07 & 93.66 & 0.0175 & 43.5094 & $+2.42\%$ &  5.00 \\
        DER &  73.26 & 93.16 & 0.0159 & 46.5341 & $+1.70\%$ & 6.00 \\
        SPG & 57.99 & 85.69 & 0.0155 & 44.7024 & $-4.55\%$ & 6.00 \\
        SGP &  65.53 & 88.44 & 0.0157 & 43.8087 & $-0.61\%$ & 4.50 \\
        \midrule
        HAD &  76.52 &  93.81 & 0.0165 & 44.2749 & $\mathbf{+3.65\%}$ & \textbf{3.50} \\
        \bottomrule
    \end{tabular}
}
\label{app_tab:city}
\vskip 0.1in
\end{table*}

\begin{table*}[h]
\centering
\caption{Performance on 3 tasks (\textit{i.e.}, 13-class semantic segmentation, depth estimation, and surface normal prediction) after the last training phase of the \textit{NYUv2} dataset compared with feature distillation methods. The best results are shown in \textbf{bold}. $\uparrow$($\downarrow$) means that the higher (lower) the value, the better the performance.}
\vskip -0.1in
\resizebox{\linewidth}{!}{
    \begin{tabular}{lccccccccccc}
        \toprule
         \multirow{4}{*}{\textbf{Method}} & \multicolumn{2}{c}{\textbf{Segmentation}} & \multicolumn{2}{c}{\textbf{Depth}} & \multicolumn{5}{c}{\textbf{Surface Normal}} & \multirow{4}{*}{\textbf{$\Delta_{v}^m \uparrow$}} & \multirow{4}{*}{\textbf{MR$\downarrow$}}\\
        \cmidrule(r){2-3} \cmidrule(r){4-5} \cmidrule(r){6-10} & \multirow{2.5}{*}{\textbf{mIoU${\uparrow}$}} &  \multirow{2.5}{*}{\textbf{Pix Acc$\uparrow$}} &  \multirow{2.5}{*}{\textbf{Abs Err $\downarrow$}} &  \multirow{2.5}{*}{\textbf{Rel Err$\downarrow$}} & \multicolumn{2}{c}{\textbf{Angle Distance}} & \multicolumn{3}{c}{\textbf{Within $t^{\circ}$}} \\ 
        \cmidrule(r){6-7} \cmidrule(r){8-10} & & & & & \textbf{Mean $\downarrow$} & \textbf{Median $\downarrow$}  & \textbf{11.25 $\uparrow$} & \textbf{22.5 $\uparrow$} & \textbf{30 $\uparrow$}  \\
        \midrule
        
         Vanilla training & 17.49 & 46.81 & 0.9609 & 0.3328 & 32.45 & 26.92 & 20.72 & 42.56 & 54.73 & $+0.00\%$ & 5.67\\
         Joint training & 41.84 & 66.14 & 0.5793 & 0.2201 & 31.53 & 25.78 & 22.38 & 44.54 & 56.36 & $+40.83\%$ & 1.00\\
        \cmidrule(r){1-11}
         Local POD w. fro  & 32.25 & 58.11 & 0.8651 & 0.3063 & 35.43 & 30.40 & 16.15 & 37.23 & 49.36 & $+25.51\%$ & 3.33 \\ 
         Local POD w. $L_1$  & 32.56 & 58.70 & 0.9240 & 0.3289 & 36.70 & 32.22 & 14.72 & 34.67 & 46.62 & $+25.77\%$ & 5.00 \\
         Local POD w. $L_2$  & 32.21 & 58.06 & 0.8658 & 0.3061 & 35.56 & 30.49 & 15.78 & 37.08 & 49.22 & $+25.66\%$ & 5.00 \\
         Global POD w. fro  & 32.31 & 58.14 & 0.8698 & 0.3077 & 35.47 & 30.41 & 16.19 & 37.24 & 49.34 & $+25.42\%$ & 4.33 \\ 
         Global POD w. $L_1$  & 32.30 & 58.12 & 0.8668 & 0.3067 & 35.45 & 30.40 & 16.07 & 37.24 & 49.36 & $+25.54\%$ & 3.33 \\ 
         HAD & 35.12 & 59.63 & 0.7410 & 0.2641 & 35.32 & 30.55 & 17.23 & 37.26 & 49.12 & $\mathbf{+32.74\%}$  & \textbf{2.00}\\      
        \bottomrule
    \end{tabular}
}
\label{tab:ablation_distil}
\vskip -0.05in
\end{table*}

The task sequences are randomly selected. For the \textit{NYUv2} dataset, the sequence is: Semantic segmentation $\rightarrow$ Depth estimation $\rightarrow$ Surface normal prediction, as shown in Tab.~\ref{tab:nyuv2}. For the \textit{Taskonomy} dataset, the sequence is: Semantic segmentation (Seg.) $\rightarrow$ Depth estimation (Dep.) $\rightarrow$ Surface normal estimation (Normal) $\rightarrow$ Edge-3D detection (E.-3D) $\rightarrow$ Reshading (Res.) $\rightarrow$ Keypoint-2D detection (K.-2D) $\rightarrow$ Edge-2D detection (E.-2D) $\rightarrow$ Euclidean distance (E.D.) $\rightarrow$ Curvatures (Curv.) $\rightarrow$ Keypoint-3D detection (K.-3D), as shown in Tab.~\ref{tab_tentask}.

For all methods, we adopt the following common settings to ensure a fair comparison. The batch size is set to 64 for the \textit{CityScapes} dataset, 16 for the \textit{Taskonomy} dataset, and 48 for \textit{NYUv2} dataset.
We use the Adam optimizer with an initial learning rate of $10^{-4}$, and adopt a linear learning rate scheduler with a warmup phase, where the warmup rate is set to 0.5. Weight decay is fixed at $10^{-5}$.

In the proposed HAD method, we perform grid searches for the hyperparameters $\alpha$, $k$, and $\tau$.
Specifically, we set hyperparameters as follows: $\alpha = 3, k = 0.5, \tau=0.9$ for \textit{NYUv2} dataset on \textit{Resnet-18}, $\alpha = 20, k = 0.6, \tau=0.6$ for \textit{NYUv2} dataset on \textit{Resnet-50}, $\alpha = 100, k = 0.6, \tau=0.6$ for \textit{CityScapes} dataset on \textit{Resnet-18}, $\alpha = 1, k = 0.8, \tau=0.5$ for \textit{CityScapes} dataset on \textit{Resnet-50},
and $\alpha = 1, k = 0.5, \tau=0.9$ for \textit{Taskonomy} dataset. 
We use the task-specific loss function as the per-pixel self-distillation loss function $\mathcal{L}_{\mathrm{dis}, j}$ of each task $\mathcal{T}_j$, \textit{i.e.}, ${\mathcal{L}}_{\mathrm{dis}, j} = \mathcal{L}_j$.

All methods are implemented using the Pytorch framework, and all models are trained on RTX V100 GPUs.

\section{Additional Results}
\label{app_results}
The results for the shuffled task sequence on the \textit{NYUv2} dataset are presented in Table~\ref{tab_seq3}. Sequence 2 consists of the following order: Surface Normal Prediction $\rightarrow$ Depth Estimation $\rightarrow$ Semantic Segmentation, while Sequence 3 follows the order: Depth Estimation $\rightarrow$ Semantic Segmentation $\rightarrow$ Surface Normal Prediction. As observed, the proposed HAD method consistently outperforms the baseline approaches, further validating its effectiveness, regardless of the task sequence.

The results of the \textit{CityScapes} dataset using \textit{ResNet-18} with different task sequences are provided in Tab.~\ref{tab:city}.
Tab.~\ref{app_tab:city} presents the results of the proposed HAD method on the same dataset using \textit{ResNet-50}, with semantic segmentation as the first task and depth estimation as the second. 
As can be seen, the proposed HAD method outperforms baseline methods in both mitigating the performance degradation of the previous task and improving overall performance.
Note that although the \textit{Cityscapes} dataset contains only 2 tasks, the LHL4DP scenario is fundamentally different from transfer learning (TL), as LHL4DP treats all tasks equally by preserving the performance of previous tasks. In contrast, TL primarily focuses on optimizing the performance of the target task.

Additionally, in Fig.~\ref{app_fig_dynamics}, we present the results of average relative improvement ($\Delta_v^m$) and mean rank (\textbf{MR}), detailed in Sec.~\ref{sec:exp_setting}, across various methods in each training phase on the \textit{Taskonomy} dataset.
As can be seen, the proposed method HAD consistently outperformed the baseline methods.

\section{Additional Ablation}
\label{app_ablation} 

To further demonstrate the effectiveness of HAD against feature distillation baselines, we compare it with the Local POD and Global POD feature distillation strategies adopted in PLOP~\cite{dong2023heterogeneous} on the \textit{NYUv2} dataset. 
As shown in Tab.~\ref{tab:ablation_distil}, all POD-based variants consistently outperform vanilla training, indicating that feature distillation is beneficial in the LHL4DP setting. 
Nevertheless, they remain inferior to HAD, which demonstrates the superiority of the proposed method in heterogeneous lifelong dense prediction tasks.

\begin{table}[h]
 \centering
    \caption{$\Delta_{b}^T$ of different hyperparameter $k,\tau$.}
    \label{tab_ab_hyper_two}
        \begin{tabular}{c|ccccc}
        \toprule
        \diagbox{$k$}{$\tau$} & 0.5 & 0.6 & 0.7 & 0.8 & 0.9 \\
        \midrule
        0.5 & -4.87 & -1.07 & -2.13 & -0.92 & 0.00 \\
        0.6 & -2.61 & -2.66 & -0.12 & -1.76 & -0.47 \\
        0.7 & -3.58 & -3.48 & -1.21 & -1.77 & -0.81 \\
        0.8 & -4.07 & -2.41 & -2.11 & -2.11 & -1.48 \\
        0.9 & -5.25 & -3.87 & -2.44 & -2.28 & -2.31 \\
        \bottomrule
        \end{tabular}
\end{table}

\begin{table}[h]
\caption{$\Delta_{b}^T$ of different hyperparameter $\alpha$.}
\label{tab_ab_hyper_alpha}
\centering
\resizebox{0.9\linewidth}{!}{
    \begin{tabular}{c|cccccc}
        \toprule
        $\alpha$ & 1 & 2 & 3 & 4 & 5 & 6 \\
        \midrule
         $\Delta_{b}^T \uparrow$ & -5.79 & -2.71 & 0.00 & -0.60 & -1.10 & -0.83 \\
        \bottomrule    
    \end{tabular}
}
\end{table}

\section{Hyperparameter Sensitivity Analysis}
\label{app_hyper_ablation}

In this section, we investigate the impact of hyperparameters on the performance of the proposed HAD method, evaluated on the $NYUv2$ dataset. Specifically, we explore the effects of the distillation weight ($\alpha$), the region partition threshold ($k$), and the Sobel threshold ($\tau$).
The results for varying distillation weights ($\alpha$) are provided in Tab.~\ref{tab_ab_hyper_alpha}, while Tab.~\ref{tab_ab_hyper_two} presents the performance variations for different values of $\tau$ and $k$.
The results presented in both tables illustrate the performance differences relative to the results of HAD reported in Tab.~\ref{tab:nyuv2}. 

The performance of the proposed HAD method remains consistently stable across a reasonable range of hyperparameter values. For example, the proposed method can achieve satisfactory results with k in 0.5-0.6 and $\tau$ in 0.7-0.9.

\end{document}